%% file: ICCV_main.tex
\documentclass[10pt,twocolumn,letterpaper]{article}

\usepackage{iccv}
\usepackage{times}
\usepackage{epsfig}
\usepackage{graphicx}
\usepackage{amsmath}
\usepackage{amssymb}
\usepackage{booktabs}
\usepackage{comment}
\usepackage{url}
\usepackage{multirow}
\usepackage{subfiles}
\usepackage{makecell}
\usepackage{bbm}
\usepackage{mathtools}
\usepackage[dvipsnames]{xcolor}
\usepackage[font=small, labelsep=period, skip=2mm]{caption}
\usepackage[accsupp]{axessibility}

\usepackage{array}
\newcolumntype{L}[1]{>{\raggedright\let\newline\\\arraybackslash\hspace{0pt}}m{#1}}
\newcolumntype{C}[1]{>{\centering\let\newline\\\arraybackslash\hspace{0pt}}m{#1}}
\newcolumntype{R}[1]{>{\raggedleft\let\newline\\\arraybackslash\hspace{0pt}}m{#1}}

\usepackage[pagebackref,breaklinks,colorlinks,citecolor=blue,linkcolor=blue,bookmarks=false]{hyperref}

\iccvfinalcopy % *** Uncomment this line for the final submission

\def\iccvPaperID{6580} % *** Enter the ICCV Paper ID here

\ificcvfinal\fi

\usepackage[capitalize]{cleveref}
\crefname{section}{Sec.}{Secs.}
\Crefname{section}{Section}{Sections}
\Crefname{table}{Table}{Tables}
\crefname{table}{Tab.}{Tabs.}

\input{def}

\begin{document}

%%%%%%%%% TITLE
\title{TextManiA: Enriching Visual Feature by Text-driven Manifold Augmentation}

\author{
% \textbf{Anonymous Author(s)}\\\\
    Moon Ye-Bin${}^{1}$ \qquad
    Jisoo Kim${}^{2}$ \qquad
    Hongyeob Kim${}^{3}$ \qquad
    Kilho Son${}^{4}$ \qquad
    Tae-Hyun Oh${}^{1,5,6}$
    % \thanks{
    % \footnotesize 
    % }
    \vspace{3mm} \\ % AO: I added vspace 
    ${}^{1}$Dept.~of Electrical Engineering and ${}^{5}$Grad.~School of Artificial Intelligence, POSTECH\\
    ${}^{2}$Columbia University \qquad
    ${}^{3}$Sungkyunkwan University \qquad
    ${}^{4}$Microsoft Azure\\
    ${}^{6}$Institute for Convergence Research and Education in Advanced Technology, Yonsei University\vspace{1mm}\\
    {\normalsize\url{https://textmania.github.io/}
    }
% Moon Ye-Bin\\
% Dept. of EE, POSTECH\\
% {\tt\small ybmoon@postech.ac.kr}
% \and
% Jisoo Kim\\
% Columbia University\\
% % First line of institution2 address\\
% {\tt\small jk4739@columbia.edu}
% \and
% Hongyeob Kim\\
% Sungkyunkwan University\\
% % First line of institution2 address\\
% {\tt\small heimish.kyma@gmail.com}
% \and
% Kilho Son\\
% Microsoft Azure\\
% {\tt\small kilho.son@gmail.com}
% \and
% Tae-Hyun Oh\\
% Dept. of EE and GSAI, POSTECH\\
% {\tt\small taehyun@postech.ac.kr}
}

\maketitle
\ificcvfinal\thispagestyle{empty}\fi

%%%%%%%%% ABSTRACT
\begin{abstract}\label{sec:0}
    \input{sections/0.abstract}
\end{abstract}

%%%%%%%%% BODY TEXT
\section{Introduction}\label{sec:1}
\input{sections/1.intro}

%------------------------------------------------------------------------
\section{Related Work}\label{sec:2}
\input{sections/2.rw}

%------------------------------------------------------------------------
\section{TextManiA}\label{sec:3}
\input{sections/3.method}

%------------------------------------------------------------------------
\section{Experiments}\label{sec:4}
\input{sections/4.experiments}

%------------------------------------------------------------------------
\section{Conclusion}\label{sec:5}
\input{sections/5.conclusion}

{\small
\bibliographystyle{ieee_fullname}
\bibliography{egbib}
}

\end{document}

%% file: def.tex
\usepackage{multirow}
\usepackage{wrapfig}
\usepackage{colortbl}
\usepackage{enumitem}
\usepackage{microtype}
\usepackage[hang,flushmargin]{footmisc}

\setlist[itemize]{align=parleft,left=0pt,topsep=1mm,itemsep=0mm,parsep=1mm}

\definecolor{azure(colorwheel)}{rgb}{0.0, 0.5, 1.0}
\definecolor{nicegreen}{rgb}{0.0, 0.7, 0.1}
\definecolor{CuGray}{gray}{0.9}
\definecolor{amethyst}{rgb}{0.6, 0.4, 0.8}
\definecolor{black}{rgb}{0.0, 0.0, 0.0}
\definecolor{steelblue}{rgb}{0.27, 0.51, 0.7}
\definecolor{brightcerulean}{rgb}{0.11, 0.67, 0.84}
\definecolor{postechred}{rgb}{0.784, 0.003, 0.313}

\newcolumntype{g}{>{\columncolor{CuGray}}c}
\newcolumntype{z}{>{\columncolor{CuGray}}l}

\renewcommand{\paragraph}[1]{\noindent\textbf{#1.}\,\,}

\newcommand{\orange}[1]{\textcolor{YellowOrange}{#1}}

\newcommand{\blue}[1]{\textcolor{Cerulean}{#1}}

\newcommand{\moon}[1]{\textcolor{black}{#1}}

\usepackage{xspace}

\def\onedot{.\@\xspace}
\def\eg{\emph{e.g}\onedot} 
\def\ie{\emph{i.e}\onedot} 
 
\def\etc{\emph{etc}\onedot} 
 
\def\etal{\emph{et al}\onedot}

\newcommand{\Sref}[1]{Sec.~\ref{#1}}
\newcommand{\Eref}[1]{Eq.~(\ref{#1})}
\newcommand{\Fref}[1]{Fig.~\ref{#1}}
\newcommand{\Tref}[1]{Table~\ref{#1}}

\def\TextMani{\texttt{TextManiA}}

\newcommand{\bolde}{{\mathbf{e}}}
\newcommand{\boldf}{{\mathbf{f}}}

\newcommand{\bz}{{\mathbf{z}}}

\newcommand{\bI}{\mathbf{I}}

\newcommand{\be}{\begin{eqnarray}}
\newcommand{\ee}{\end{eqnarray}}
\newcommand{\bee}{\begin{eqnarray*}}
\newcommand{\eee}{\end{eqnarray*}}

\newcommand{\matrixb}{\left[ \begin{array}}
\newcommand{\matrixe}{\end{array} \right]}

%% file: sections/0.abstract.tex
We propose $\TextMani$, a text-driven manifold augmentation method that semantically enriches visual feature spaces, regardless of class distribution.
$\TextMani$ augments visual data with intra-class semantic perturbation by exploiting easy-to-understand visually mimetic words, i.e., attributes.
This work is built on an interesting hypothesis that general language models, \eg, BERT and GPT, encompass visual information to some extent, even without training on visual training data.
Given the hypothesis, $\TextMani$ transfers pre-trained text representation obtained from a well-established large language encoder to a target visual feature space being learned.
Our extensive analysis hints that the language encoder indeed encompasses visual information at least useful to augment visual representation.
Our experiments demonstrate that $\TextMani$ is particularly powerful in scarce samples with class imbalance as well as even distribution.
We also show compatibility with the label mix-based approaches in evenly distributed scarce data.

%% file: sections/1.intro.tex
% Neural networks have been a strong backbone of many industries as well as research fields. One of crucial success-requirements of the \emph{data-driven} neural networks is \emph{relevant} and \emph{sufficient} data. Neural networks normally degenerate their performance on target tasks if the train and test data present even slightly different distributions~\cite{ben2010theory,szegedy2014intriguing,verma2019manifold}. 
% Neural networks often present high accuracy on train data, whereas suffer from under-performance on test set
% if data is scarce. 
% In practice, acquiring data
% with those properties is uneasy or expensive.
% Medical data is expensive and uneasy to acquire but synthetic or other data can not be a complete replacement due to the domain shift.
% In open world recognition, where the target classes
% % concepts (or classes) 
% are almost unlimited, collecting sufficient data with ground-truth on all the classes is almost impossible.

% Deep neural networks have been the backbone model of the state-of-the-art systems and have enabled breakthroughs in diverse fields including computer vision~\cite{lecun2015deep}.

% Neural networks 
Learning models, \eg, neural networks, are known to perform well on visual recognition tasks when training and testing datasets 
% that
present similar distributions~\cite{ben2010theory}.
However, their performance often degrades considerably when evaluated in subtly different distributions
% The performance, however, often degrades considerably when evaluated in subtly different distributions
% shifted even slightly
\cite{szegedy2014intriguing}.
One effective way to enhance the generalization ability of 
% make
a model 
% generalized 
against such data distribution shifts would be
% is
data augmentation~\cite{devries2017improved,zhang2017mixup,yun2019cutmix,li2021simple,li2021feature,verma2019manifold}.
Augmenting data enlarges the support of the training distribution formed
% defined
by given samples and yields the effect of increasing the amount of data even without additional laborious data collection.
By training on augmented data, decision boundaries 
% \oh{can secure margins,}
are smoothed, 
and the generalization ability of the model is improved~\cite{verma2019manifold}.
% ability is increased

\begin{figure}
    \centering
    \vspace{2mm}
    \includegraphics[width=1.0\linewidth]{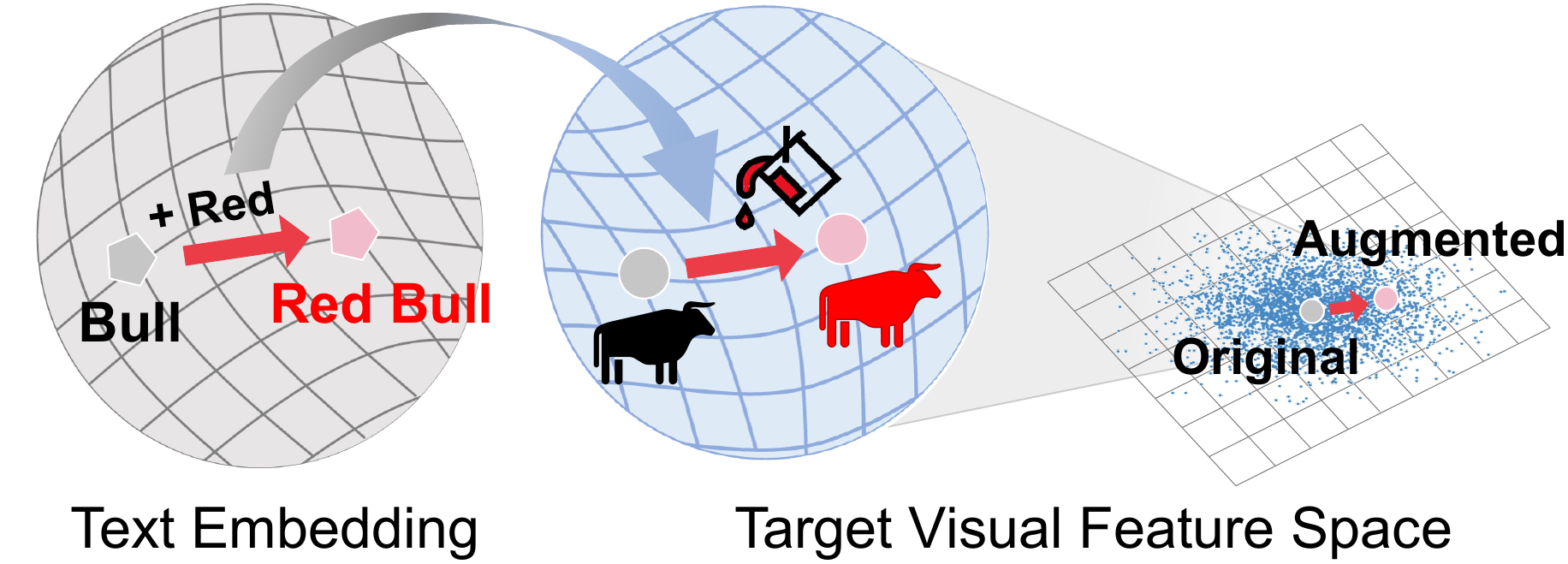}\vspace{2mm}
    \caption{Illustration of $\TextMani$.
    Our method augments the target visual feature by leveraging text embedding of the visually mimetic words, which are comprehensible and semantically
    % physically
    rich.
    For example, when the text of the existing class ``bull'' is manipulated as ``red bull'' by adding the attribute ``red,'' we can get augmented visual features by reflecting the difference of text embeddings.
    % to the visual feature space.
    In this way, $\TextMani$ densifies sparse visual feature space using various attributes text.
    }
    \label{fig:teaser2}
\end{figure}

% Visual data augmentation should need careful design founded on understanding visual data generation process, models, and target tasks.
There has been a distinctive and successful line of research for label mix-based data augmentation, such as Mixup~\cite{zhang2017mixup}, CutMix~\cite{yun2019cutmix}, and manifold Mixup~\cite{verma2019manifold}, which are
% is
effective for model generalization and calibration~\cite{guo2017calibration}.
% There are simple but effective mix-based visual data augmentations, such as Mixup~\cite{zhang2017mixup}, CutMix~\cite{yun2019cutmix}, and manifold Mixup~\cite{verma2019manifold}.
The effectiveness of those label mix-based approaches is attributed to semantic perturbation by label mixing~\cite{zhang2017mixup,verma2019manifold,yun2019cutmix}.
This is a distinctive property from other lines of data augmentation methods, \eg, \cite{vincent2010stacked,li2021simple,li2021feature,simard1998transformation,cubuk2020randaugment}, where they synthesize diverse virtual data that appear differently but retain class semantics of original contents.
% They augment data by interpolating the input images or those features as well as mixing those labels. 
% with the mixing ratio corresponding to the input mix.
% which additionally have a model calibration effect.
However, we found that the performance of mix-based augmentation methods is
% are
noticeably degraded when 
% applied to
training with 
% the effect of mix-based augmentation is reduced when we have only 
skewed class distribution having scarce samples for non-major classes, \ie, long-tailed distribution.
% ~\cite{yang2022survey,cui2019class,cao2019learning}.
In 
% the
real-world, data often exhibit long-tailed class distribution (\eg, 
% e.g.,
Pareto distribution), which cannot be dealt with the prevalent mix-based approaches.
% Frustratingly, real-world data often exhibit long-tailed class distribution.
This motivates us to seek a semantically rich data augmentation 
% that is 
effective for limited data regimes, including long-tailed distribution, scarce data, and few-shot cases.

In this work, we propose $\TextMani$, a text-driven manifold augmentation for visual features, which is effective for 
% takes account of
long-tail classes and scarce data.
% \oh{Our hypothesis 
% We hypothesize that 
% visual attribute words represented by general language models, \eg, BERT~\cite{devlin2018bert} and GPT~\cite{radford2019language}, may already be learned to encode some visual concept information to some extent, which may be transferred to the visual embedding space.
Our $\TextMani$ is based on an interesting hypothesis
% We hypothesize 
that general language models, \eg, BERT~\cite{devlin2018bert} and GPT~\cite{radford2019language}, have learned visual information to some extent that can be transferred to visual feature spaces even with no visual training data.
% With this hypothesis, we develop our method enriching the visual feature space in a semantically meaningful way by leveraging visually mimetic text.
% \son{With this hypothesis, we leverage visually mimetic texts, encoded with general language models and transferred to target visual feature space, to semantically enrich the target space.}
With this hypothesis, we semantically enrich the target visual feature space to be trained by leveraging visually mimetic texts, encoded with general language models and transferred to the target space.
% in a semantically rich meaningful way.
% Based on the hypothesis that the relationship between visual attributes in general language models would be related to the visual embedding space, our method enriches the visual feature space in a semantically meaningful way
% % latent augmentation method 
% by leveraging visually mimetic text.
% , called $\TextMani$, to tackle limited data challenges.
% including long-tailed data, few-shot, and scarce data problems.
% $\TextMani$ is designed to mimic attribute changes of input on the feature space by adding meaningful attribute vectors to input features.
Specifically, $\TextMani$ encodes meaningful attributes such as ``red'' and ``large'' to vectors by computing the difference between text embeddings 
% of text 
with and without attributes.
We add the attribute embeddings to target visual features to mimic those attributes on the target visual feature space.
% \son{
% Specifically, $\TextMani$ encodes meaningful attribute texts such as "red" and "large" to vectors with general language models and adds them to target visual features to mimic those attributes on the target feature space. 
% Due to the difference between language feature spaces and the target visual feature spaces, we transform gradient of vectors: difference between text embeddings with vs without attributes, before adding them to the target feature space.
% }
%Figure~\ref{fig:teaser2} illustrates the augmentation process of $\TextMani$ on feature level,
% (latent representation), 
% where
% Specifically, $\TextMani$ is illustrated in .
%the input feature (\eg, the visual feature of ``bull'') is manipulated by adding the attribute vector induced by the attribute text (\eg, ``red''), which yields the augmented visual feature (\eg, ``red bull'').
Figure~\ref{fig:teaser2} illustrates the augmentation process of $\TextMani$. 
The input feature (\eg, the visual feature of ``bull'') is manipulated by adding the attribute vector induced by the attribute text (\eg, ``red''), which yields the augmented visual feature (\eg, ``red bull'').
% mimics attribute changes of input
% on the feature space (\eg, to the visual feature of ``red bull'').
% We note that $\TextMani$ uses a controlled text. 
Thanks to the text modality properties, the augmentations generated by $\TextMani$ are symbolic, human-interpretable, 
% physically relevant 
and easily controllable.
% We induce attribute vectors for augmentation from texts due to its modality benefits that are symbolic, human interpretable, physically relevant and easily controllable by its symbolic property.
% \moon{For example, $\TextMani$ uses modifier words, such as ``red,'' ``blue,'' ``large,'' and ``small,''  and returns the augmented visual feature related to the words.}
% For example, modifier words such as ``red'', ``blue'', ``large'' and ``small'' are used to control $\TextMani$ and it returns augmentations visually related to the words.

Our approach applies semantic perturbation on a different level to that of the label mix-based methods~\cite{zhang2017mixup,verma2019manifold,yun2019cutmix}.
The mix-based methods augment a sample from a combination of two different class samples, \ie, applying semantic perturbation in an \emph{inter-class} way.
% by mixing two different source classes.
This further aggravates the class imbalance problem in the long-tailed (skewed) class distribution cases.\footnote{For example, if data size of major classes is 10 times larger than 
% one in 
minor classes, the probability of choosing a pair of source samples from the major classes is approximately 100 times more than that of 
% ones from
minor classes.}
% In contrast, we perturb in an \emph{intra-class} way which does not alter the class label of a sample but enriches semantic granularity of the class.
Our $\TextMani$, whereas, perturbs data in an \emph{intra-class} way. 
A sample per each class is selected, and we enrich the semantic granularity of the class using the sample, thus enabling us to better maintain the amount of augmentation balances in the long-tailed class distribution cases.
Moreover, $\TextMani$ can densify around the training samples by extrapolating the class semantics along augmented semantic attribute axes.
With this, our method 
% further,
can be combined with the label mix-based methods to further improve performance in evenly distributed sparse data cases because they are complementary.

\moon{To empirically support that our attribute vectors transformed from
% estimation with 
text embeddings are reasonably designed, we devise two visualization-based analyses: with t-SNE~\cite{van2008visualizing} and a latent inversion technique.}
\moon{These demonstrate that attribute vectors lead to visually interpretable manifold augmentation of input.}
% images.
% , which validates our design.
% These results show that the difference vector between the original and variant text embeddings embeds visual attribute information, and we may use it to inject the attribute signals to the target feature space.
We also evaluate our method with two different tasks in scarce data regimes: few-shot object detection and image classification with deficient datasets and long-tail datasets. 
% The considerable 
Our experiments demonstrate that $\TextMani$ is an effective and model-agnostic data augmentation method, especially in 
% the
scarce data cases, by exploiting the favors of zero-shot attributes.
Also, additional studies show the versatility and compatibility of the design of $\TextMani$. 
Our key contributions are summarized as:
% \vspace{-1mm}
\begin{itemize}
    \item We propose $\TextMani$, which enriches the visual features
    % is a generic zero-shot feature augmentation 
    % for densifying sparse samples 
    by conveying \moon{attribute information from the text embedding to the target visual feature space.}
    % text attribute information to the target feature space.
    % The residual information is from the class name as text input and its variant with attribute words, which are explainable and controllable.
    % \vspace{-1.5mm}
    % \item We devise two visualization-based analyses to empirically support that our attribute vector estimation with text embedding is reasonably designed.
    \item We validate our hypothesis of the existence of embedded visual knowledge in pre-trained language encoders despite no training on visual data.
    \item We demonstrate that $\TextMani$ is especially helpful in augmenting sparse samples in long-tail class cases.
    \item We show that our $\TextMani$ is complementary to other augmentation methods, and in particular, the combination of our $\TextMani$ and manifold Mixup~\cite{verma2019manifold} noticeably improves the performance in deficient data cases.
    % We demonstrate adding attribute words produces scattered embedding vectors around the original points by visualizing the CLIP text embedding. 
    % We also give evidence that we can utilize the difference vector to transfer the attribute signal to the target feature space through simple image manipulation examples.
\end{itemize}

%% file: sections/2.rw.tex
We brief the related work in the following three perspectives:
% from the perspective of methods, \ie, 
image data augmentation, foundation models, and target application tasks.
In this work, our $\TextMani$ augments data by leveraging 
% using
% text features obtained from 
the text encoder of CLIP~\cite{radford2021learning}, BERT~\cite{devlin2018bert}, or GPT-2\footnote{GTP-2 is the decoder-only architecture, but we use it as a text embedding extractor, so we call it text-encoder.}~\cite{radford2019language}.
% , and whereby we tackle data scarce regimes including
% To validate that $\TextMani$ is model-agnostic and suited for scarce data cases, we evaluate on 
% long-tail class distribution, small dataset, and few-shot problems.
For main target applications, 
% mainly dealt within this work, 
we focus on long-tail and small data classification and few-shot object detection tasks in the data-scarce regimes.

\paragraph{Image Data Augmentation}
% \cite{simard1998transformation}
% Data augmentation is the simple and effective technique that makes models train on similar but different augmented samples to given training samples.
% These augmented samples can be obtained by applying perturbation to the original training samples, which enlarges the support of the training distribution defined by samples and produces the effect of increasing the amount of data. 
% From the viewpoint of an embedding space, data augmentation densifies
% % can fill the 
% empty spaces among 
% % between the 
% original data points without 
% % the
% additional laborious data collection.
% It is beneficial in practice, where only sparse data samples are available.
% This has shown to be surprisingly effective to generalization. 
% augment the amount of data by applying various transformations to 
% the
% original images.
% Through augmented data, a model can be more generalized and improve performance.
% Image data augmentation can be largely divided into whether semantic perturbation exists along label mixing.
Image data augmentation can be largely divided into whether semantic perturbation exists.
Semantic perturbation, in specific, can be further split into methods with or without label mixing.
% image-level and feature-level augmentations.
% In image data augmentation methods, basic perturbations applied to original training images are 
Methods~\cite{devries2017improved,srivastava2014dropout,reed1999neural,an1996effects, gulcehre2016noisy, bishop1995training, holmstrom1992using, vincent2010stacked,li2021simple,alemi2016deep,li2021feature,simard1998transformation,cubuk2020randaugment,cubuk2018autoaugment} without semantic perturbation, which have no label change, contain
% The image-level augmentation directly manipulates a training image without changing its label, 
% such as 
primitive image processing and transformation operations.
This includes photometric (\eg, color jitter, contrast, blur, noise, \etc) and geometric (\eg, horizontal reflection, rotation, \etc) operations, and
% color jitter, blur, 
% add
% noise, 
% random
% cropping, flipping, rotation, 
advanced augmentations, including Cutout~\cite{devries2017improved} and adaptive combinations~\cite{cubuk2020randaugment,cubuk2018autoaugment}.
% RandAugment~\cite{cubuk2020randaugment}, and AutoAugment~\cite{cubuk2018autoaugment}, 
% or with label manipulation, such as 

In contrast, Mixup~\cite{zhang2017mixup}, CutMix~\cite{yun2019cutmix}, and manifold Mixup~\cite{verma2019manifold} execute semantic perturbation along with label mixing.
Mixup interpolates two whole input images pixel-wisely, 
% in a pixel level,
% within the overall region of an image, 
CutMix interpolates a partial region of an image with another, and manifold Mixup mixes features from the images.
These mix-based methods also augment labels of samples by an inter-class semantic perturbation, where labels of two different class samples are mixed. 
While the mixed label is known to be effective for generalization and model calibration effects~\cite{guo2017calibration}, we found that 
% \moon{which would be consistent with Cha~\etal~\cite{cha2021swad}.}
% They include the process of sampling two targets to be mixed from the dataset.
% If the dataset with a non-uniform distribution has long-tail classes, the probability of being sampled between classes would be different, which can degrade the performance.
the mix-based methods are heavily affected by class distribution due to sampling from two sources; thus, their effect is restricted to evenly distributed datasets.
For datasets with skewed
% non-uniform 
class distributions with tails,
% classes, 
the sampling probabilities between major and minor classes would significantly differ, which can exaggerate biased sampling to major classes and makes minor classes more minor. \\
% can lead to accuracy degradation.
% of the performance.

Our $\TextMani$, on the other hand, is applied to all of the given samples uniformly regardless of class distribution.
$\TextMani$ densifies around the sample features by perturbing and enriching the semantic meaning of them
% the samples 
at an intra-class level, which does not change the label.
Moreover, because of the different semantic granularity of perturbation between $\TextMani$ (intra-class) and mix-based methods (inter-class), two methods can be used complementarily when class imbalance does not exist.

\paragraph{Foundation Models}
Recent foundation models~\cite{yuan2021florence, radford2021learning, li2022grounded, jia2021scaling, ramesh2021zero, devlin2018bert, radford2019language} have shown a successful case of reflecting human nuances with visually imitated word composition.
% Especially, 
Particularly, language models, \eg, BERT~\cite{devlin2018bert} and GPT~\cite{radford2019language}, show their ability not only in language tasks~\cite{wu2020tod} but also in vision-language multi-modal tasks~\cite{sammani2022nlx,gal2022image}.
Contrastive Language-Image Pretraining (CLIP)~\cite{radford2021learning} also achieves huge success in various tasks even in zero-shot recognition.
% CLIP contains the image and text encoders and is trained with contrastive loss on the large-scale image and text pairs.
% The multi-modal joint embedding space of the CLIP is learned for zero-shot image and language-related recognition tasks.
Follow-up studies show that CLIP representation is effective in conducting other visual tasks by bridging vision and language, \eg, 2D image generation~\cite{gal2022stylegan, kwon2022clipstyler, kim2022diffusionclip, nichol2021glide}, image manipulation~\cite{patashnik2021styleclip, kim2022diffusionclip} and synthesis~\cite{frans2021clipdraw}, and even 3D domain tasks~\cite{youwang2022clip, jain2022zero, michel2022text2mesh}.
% Prior works optimize the features with CLIP loss to output manipulated images, pose or textures.

In $\TextMani$, we focus on estimating attribute features by exploiting BERT, GPT-2, or CLIP text encoder alone. 
Distinctively, we only transfer the estimated attribute feature to augment visual features in a different space, which makes our work different from knowledge distillation~\cite{hinton2015distilling} of foundation models~\cite{dai2022enabling,wang2022multimodal,shin2022namedmask}.
Rather, our design is an instance of the module neural network structure~\cite{happel1994design,andreas2016neural}, where recent module-based designs procedurally train the whole model module-by-module with the guidance of the well pre-trained module, \eg, \cite{kumar2021rma,rombach2022high,oh2019speech2face,sung2023sound,gupta2023visual}.
Also, our work is applicable agnostically to architectures; thus, 
% which is
more flexibly applicable than fine-tuning of foundation models~\cite{wortsman2022robust}.

% $\TextMani$, however, focused on feature manipulation exploiting CLIP embedding to extract the residual information and improve target task performance.

\paragraph{Long-tail Classification}
In real world, visual data follow a long-tailed distribution which induces class imbalance and leads to
% Model performance heavily relies
% has a heavy reliance 
% on data quality, which is affected by the imbalance factor~\cite{yang2022survey};
% thus, model 
performance degrading~\cite{yang2022survey}.
% on heavy tail data distribution.
A representative line of the methods for long-tail classification is rebalancing~\cite{buda2018systematic, cui2019class, ren2018learning}, which resamples data or reweights the loss for tail classes.
However, improvement in performance of the tail classes comes with the sacrifice of head class performance.
Note that $\TextMani$ densifies all the given samples regardless of the class imbalance, and whereby 
% So 
the model is trained with reasonable variations of training samples for every class at least,
% has a lower bound in number of samples for every class, 
which improves the performance while minimizing sacrifice of the head class.
% Note that $\TextMani$ densifies all the given samples regardless of the class imbalance so the model can see a certain number of samples for all classes, which improves the performance with minimum sacrifice of the head class.

% The prior work for long-tail classification can be classified into class rebalancing~\cite{buda2018systematic, cui2019class, ren2018learning}, multi-stage training~\cite{cao2019learning, kang2019decoupling}, and multi-expert~\cite{cai2021ace, xiang2020learning, zhou2020bbn, wang2020long} methods.
% Class rebalancing methods resample the data or reweight the loss for the tail classes, which improves the performance of the tail classes with the sacrifice of the performance of the head classes.
% Multi-stage training methods consist of representation learning and classifier training stages to learn richer representations that can encode even tail distribution.
% Multi-expert methods engage multiple models to learn different aspects of information and ensemble or distill the knowledge of each other.
% Note that $\TextMani$ augments all data so that the model can see a certain number of samples for all classes.
% $\TextMani$ could be used in combination with existing long-tail classification methods because it is orthogonal with existing methods.

\paragraph{Few-Shot Object Detection (FSOD)}
% In this work, w
We 
% specifically
tackle FSOD, one of the sparse sample problems, to demonstrate the effectiveness of $\TextMani$ and its model architecture agnostic property.
% of our proposed method.
FSOD handles novel object classes after the base training for object detection tasks.
The model rapidly adapts to novel classes using few data by matching-based~\cite{li2021beyond, chen2021dual, xiao2020few} or fine-tuning based~\cite{hu2021dense, sun2021fsce, wang2020frustratingly, qiao2021defrcn, ye2023eninst} methods.
% Matching-based methods train the object detection model by measuring the similarity between the query objects and the given small number of novel objects.
% Fine-tuning based methods force the model to adapt its parameters fast to novel object classes with few examples.
$\TextMani$ is evaluated with the fine-tuning-based FSOD approach~\cite{wang2020frustratingly},
% task in a 
% % with
% fine-tuning approach, 
% based methods, 
which facilitates to use 
% have
general model architectures.
% Note that we apply $\TextMani$ only on the classification head; thus, the quality of the regressed bounding boxes will be the same as before applying $\TextMani$.

% There are several data augmentation methods~\cite{hariharan2017low, wang2018low} in the few-shot classification, which generates additional images of novel classes using generative models.
% They are similar to $\TextMani$ in the sense of enriching the data, but different in augmentation level and target task.
% $\TextMani$ augments the data by manipulating the feature, which is more simple than image generation.

%% file: sections/3.method.tex
\begin{figure}[t!]
\centering
\includegraphics[width=0.95\linewidth]{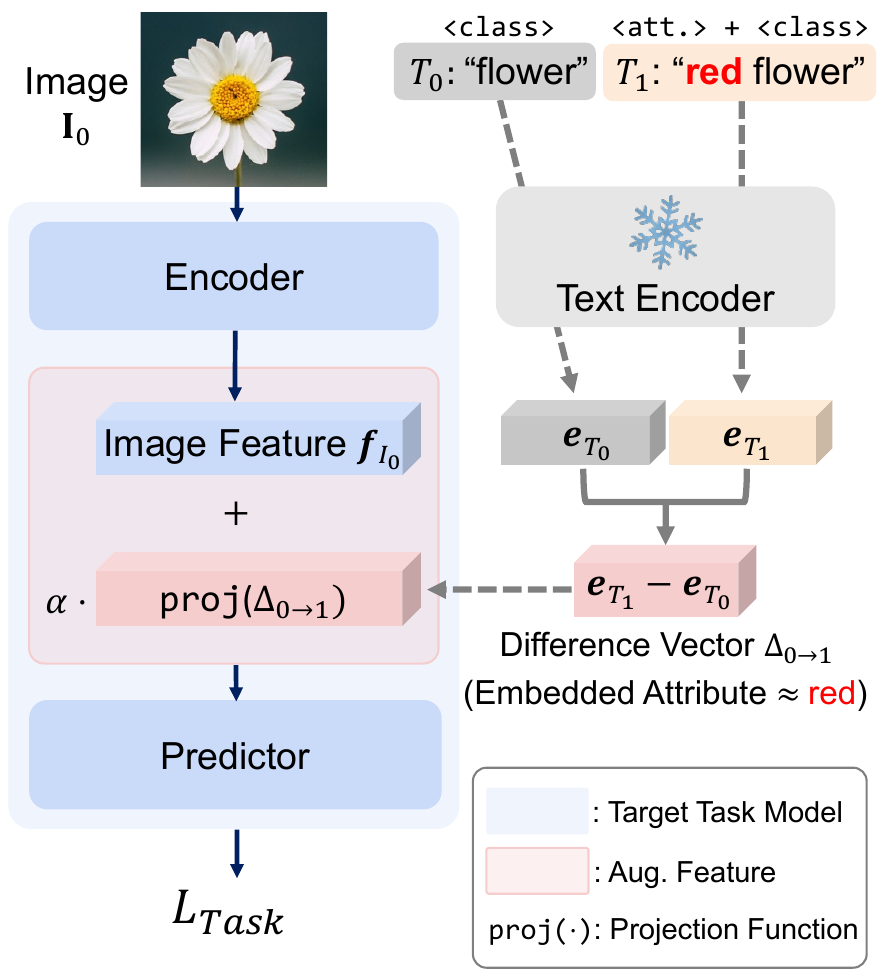}
\caption{Overview of $\TextMani$. 
Given flower image $\bI_0$ and class ``flower'' $T_0$, we construct the variant text $T_1$ by adding the attribute ``red'' on $T_0$.
% Difference vector, $\mathbf{\Delta}_{0\to1}=\bolde_{T_1} - \bolde_{T_0}$ computed with the CLIP text encoder embedding vectors of text and variant text is added on the image feature $\boldf_{I_0}$ with projection and $\alpha$ weighting.
Embeddings of $T_0$ and $T_1$ are computed with text encoders, \eg CLIP~\cite{radford2021learning}, BERT~\cite{devlin2018bert}, or GPT-2~\cite{radford2019language} and their difference vector, $\mathbf{\Delta}_{0\to1}=\bolde_{T_1} - \bolde_{T_0}$ is added to the image feature $\boldf_{I_0}$ after projection $\texttt{proj}(\cdot)$ and weight $\alpha$.
We make the target feature space semantically rich and plausible by adding the difference vector, which embeds interpretable information.
% We manipulate the zero-shot attribute to make the target feature space semantically plausible and rich by adding the difference vector, which embeds the interpretable information. 
%\son{\ie, attribute}.
% We train the target task model by augmented target visual features.
}
\label{fig:textmani}
\end{figure}

In image classification, the class label is typically utilized only as a supervision for measuring the loss. 
We, instead, propose to treat the class label as additional information, the text describing the class, and derive semantic information from it. 
% We, instead, propose to treat the class label as one more information, the text describing the class, and derive semantic information from it. 
% However, only a class label as a text description is too coarse to represent rich semantics within a class.
However, class label as a text description itself is too coarse to represent rich semantics within a class.
For example, a class label ``dog'' does not represent all the details of the description such as ``small size of the brown colored dog.''
% and the large size of the black colored dog.
To enrich
% make the model understand the
% diverse levels of 
the detailed semantics over
% from
the given coarse class texts,
% To unpack the information of compact class text and augment the sparse image data, 
we leverage the attribute words, such as ``small size'' and ``brown colored,'' that can visually modify objects in images at the semantic level.

\subsection{Main Idea}\label{sec3.1}
The main idea of $\TextMani$ is to densify distribution around sparse training samples on the target
% model's
feature space, making it semantically rich through the difference vectors having plausible attribute information,
% in a semantically meaningful way 
% through
% % by using semantically 
% plausible zero-shot attributes,
% to improve the performance of the target model,
% 
as depicted in \Fref{fig:teaser2}.

Figure~\ref{fig:textmani} illustrates how $\TextMani$ augment data.
% the way to augment data with $\TextMani$.
Suppose we have an image $\bI_0$ and 
% its
corresponding class label $T_0$.
The model generally learns the target task using the image $\bI_0$ as an input and the class label $T_0$ as supervision.
In this work, we also consider the class label $T_0$ as text information and extract the embedding vector $\bolde_{T_0} \in \mathbb{R}^{d_c}$ using text encoder, \eg, CLIP~\cite{radford2021learning}, BERT~\cite{devlin2018bert}, or GPT-2~\cite{radford2019language}, where $d_c$ is the text embedding dimension.
For obtaining an embedding vector $\bolde_{T_0}$, we use the text embedding of the encoder output directly when using CLIP text encoder, or use the average vector of 
% computed from 
all the embeddings of the sentence when using other language models such as BERT or GPT-2.
% When using BERT or GPT-2, we average all the embeddings from the sentence and use it as an embedding vector $\bolde_{T_0}$.
% In this way, we can leverage the visual-language joint embedding space to represent text as a visually relevant vector and obtain visual attributes from given texts.

Specifically, text input $T_0$ is formed with class name and pre-defined prompts, such as ``a photo of,'' ``a picture of,'' and ``a sketch of.''
We also synthesize another text input variant $T_1$ by adding color or size attribute words, such as ``red'' and ``big,'' and compute the embedding vector $\bolde_{T_1} \in \mathbb{R}^{d_c}$.
Numerous variants can be created with various attribute words and their combinations, but we explain the case of one variant for convenience.
Based on the word vector analogy\footnote{It was shown that simple algebraic operations can be performed on the word vectors, \eg, king - man + woman $\approx$ queen on the embedding space.}~\cite{mikolov2013efficient},
we hypothesize that the relationship between $T_0$ and $T_1$ is maintained in the text embedding space, \ie, 
% the difference by attribute in the text input can be regarded as the difference of feature vectors $\bolde_{T_0}$ and $\bolde_{T_1}$ in the text embedding space; 
% thus, 
the difference vector $\mathbf{\Delta}_{0\to1}=\bolde_{T_1} - \bolde_{T_0}$ would contain the information of added attributes (this hypothesis is validated 
% later
in \Sref{sec3.2}).
To exploit the difference vector from text embeddings, we design our method on the manifold.
% motivated by manifold augmentation~\cite{verma2019manifold}.

% While
We can obtain such diverse attribute vectors from various attribute text templates; however,
% samples, 
their representation space is not directly related
% completely irrelevant 
to the visual feature space of the target model we are interested in.
To bridge the gap, we project the difference vector 
% on the target dimension
%using a learnable linear projection layer $\texttt{proj}(\cdot)$ to the target feature space, 
to the target feature space with a learnable linear projection layer $\texttt{proj}(\cdot)$.
Then, we
% and 
add the projected difference vector to the target image feature $\boldf_{I_0} \in \mathbb{R}^{d_t}$ obtained from the target task encoder with the input image $\bI_0$, where $d_t$ is the target feature space dimension.
A linear layer for $\texttt{proj}(\cdot)$ would be sufficient to transfer cross-modal information, referring to 
% results
 the cross-modal transferability under the contrastive learning case
% of the multi-modal contrastive model
\cite{zhang2023diagnosing} and our experiments.
% the dimension of the target feature space.

% To exploit the attribute vector on the target feature space, we add the difference vector on the image feature  from the target task encoder, 

% When the dimension of CLIP text embedding and target feature space are the same, \ie, $d_c = d_t$, we can add the difference vector on the target image feature directly.
% When they are different, \ie, $d_c \not = d_t$, 

To inject the stochasticity, 
a mixing weight $\alpha \in \mathbb{R}$ is introduced and randomly sampled from the clamped Normal distribution in the range over $0.1$. %$[0.1, \infty)$.
Then, we have the augmented feature vector $\hat\boldf_{I_0}$ as,
\begin{equation}
    \label{eq:aug_vec}
    \begin{aligned}
    % \begin{dcases*}
        \hat\boldf_{I_0} = \boldf_{I_0} + \alpha\cdot \mathtt{proj}(\mathbf{\Delta}_{0\to1}).
        % & if $d_c \not = d_t$,\\
        % \boldf_{I_0} + \alpha\cdot \mathbf{\Delta}_{0\to1}, & if $d_c = d_t$,
    % \end{dcases*}
    \end{aligned}
\end{equation}
For the cases having $d_t = d_c$, we can set $\mathtt{proj}(\cdot)$ operation to be an identity mapping without any learnable parameter.

% where $\mathtt{proj}(\cdot)$ stands for projection layer.
We train the target task model with this augmented feature vector, whose class label is still $T_0$. %\ie, intra-class semantic perturbation.
We note that computing difference attribute vectors with text encoder is computationally expensive. 
For efficient training, we pre-compute all possible combinations of difference vectors $\{\mathbf{\Delta}\}$ and store them in a look-up table because class names and attributes can be
% texts are
pre-determined and unchanged during training.
% Note that 

Different from knowledge distillation~\cite{dai2022enabling,wang2022multimodal,shin2022namedmask,hinton2015distilling,kim2021distilling}, $\TextMani$ does not transfer-learn the text embeddings directly. 
Instead, the difference
% attribute
vector projected onto the target domain is injected into the target model, allowing our method to be applied to 
% which enables us to apply $\TextMani$ to
arbitrary target models.
Since the visual feature augmentation is solely controlled by text, $\TextMani$ is human-interpretable and easily controllable.
% tractable. 
% while hallucinating similar effects. 
% A trade-off is to use a text encoder to define a visual-language joint embedding space and to estimate visually plausible attribute vectors from texts.
% , where we use the pre-trained CLIP text encoder~\cite{radford2021learning} 

% By adding the difference vector from the CLIP text embedding, $\TextMani$ induces the margin of difference vector size around the target image feature $\boldf_{I_0}$ through the interpretable residual information and densifies all the given samples in a uniform probability.
% % the sparse distribution.
% % thus, the performance of the target task model is improved due to the margin.
% Our $\TextMani$ augments the image data at the feature-level in a simple and interpretable way.

\begin{figure}
    \centering
    \includegraphics[width=1.0\linewidth]{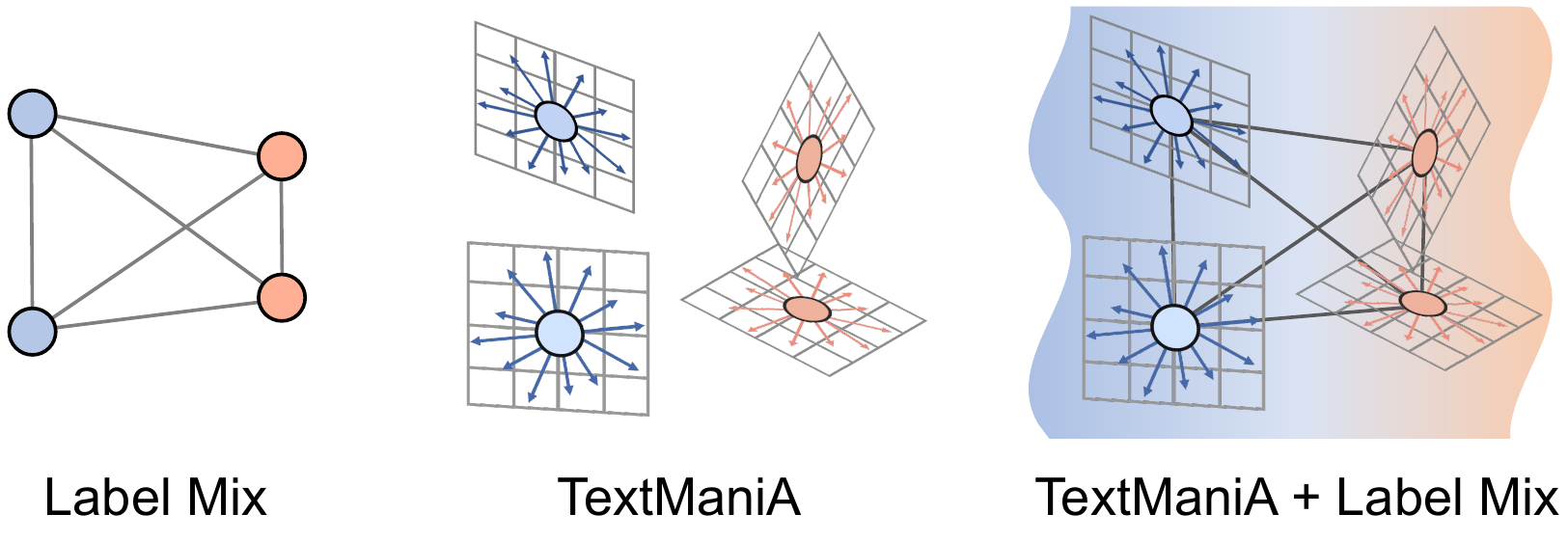}
    \vspace{1mm}
    \caption{Comparison of a typical label mix-based augmentation, $\TextMani$, and their combination, in a sparse data case.
    Given two samples for two classes, the augmented samples by a label mix-based method are at points lying on six lines resulting from the combinations of samples, \ie, inter-class perturbation.
    $\TextMani$ densifies the sample distribution along semantic attribute axes in an intra-class.
    % in specific directions.
    The combination of them yields synergy of their respective advantages.
    % When we combine them, we can exploit their respective advantages.
    }
    \label{fig:vs_mix}
\end{figure}

% In comparison 
Compared to label mix-based augmentations~\cite{zhang2017mixup,verma2019manifold,yun2019cutmix}, $\TextMani$ has advantages in imbalanced data distribution.
% As depicted in \Fref{fig:vs_mix}, 
% let us
We suppose a scenario where few samples are in one class and many samples are in another class.
% two training samples are given for each of the two tail classes.
The augmented points by a mix-based method would be located only on the interpolation lines between the given samples, which limits the augmentation effects, as depicted in \Fref{fig:vs_mix}.
If we apply a mix-based method in the long-tailed class distribution cases, \ie, notably skewed distribution, 
the class imbalance problem is further aggravated, and augmentation is more biased toward major classes.
% This is because the mix-based methods augment a sample from a combination of two different class samples, where the probability of the tail classes to be sampled is decreased by square. 
% less sampled in a batch.
% selected.
% In contrast, $\TextMani$ can equally densify all the given samples.
In contrast, $\TextMani$ can equally densify all the given samples since it augments each sample independently.
% in a uniform probability.
% applies uniform effects regardless of class imbalance.
% By adding the difference vector from the CLIP text embedding, $\TextMani$ induces the margin of difference vector size around the target image feature $\boldf_{I_0}$ through the interpretable residual information and densifies all the given samples in a uniform probability.
Thus, $\TextMani$ can be used in general regardless of the imbalance factor of class distribution.

On the other hand, 
% \Fref{fig:vs_mix} suggests
in another scenario with small training data but with uniform class distribution,
% When only small data are considered, 
% In this case, 
both $\TextMani$ and mix-based methods would increase diverse combinations of samples by augmentation in respective aspects, which leads to complementary performance improvement. 
% This may hint that combining
% % We can combine 
% the two complementary methods is favorable for mild uniform class distribution. 
This will be empirically demonstrated in \Sref{sec:4}.% cases.
% can be obtained.

\subsection{Characteristics of Attribute Embedding}\label{sec3.2}
To scrutinize the relationship between the text $T_0$, text variant $T_1$ and the attribute embedding $\mathbf{\Delta}_{0\to1}$, we visualize their distribution and discuss the characteristics.
We also visualize the difference vector to verify the hypothesis that the difference vector embeds its corresponding attribute.

% $\TextMani$ is designed upon two hypotheses: (1) the embedding vectors of variants of the class names are distributed near the original point in the text embedding, \ie, vicinal samples, and (2) the difference vector embeds the attributes.
% To verify the hypotheses, we visualize the text embedding and difference vector in this section.

\begin{figure}
    \centering
    \includegraphics[width=0.95\linewidth]{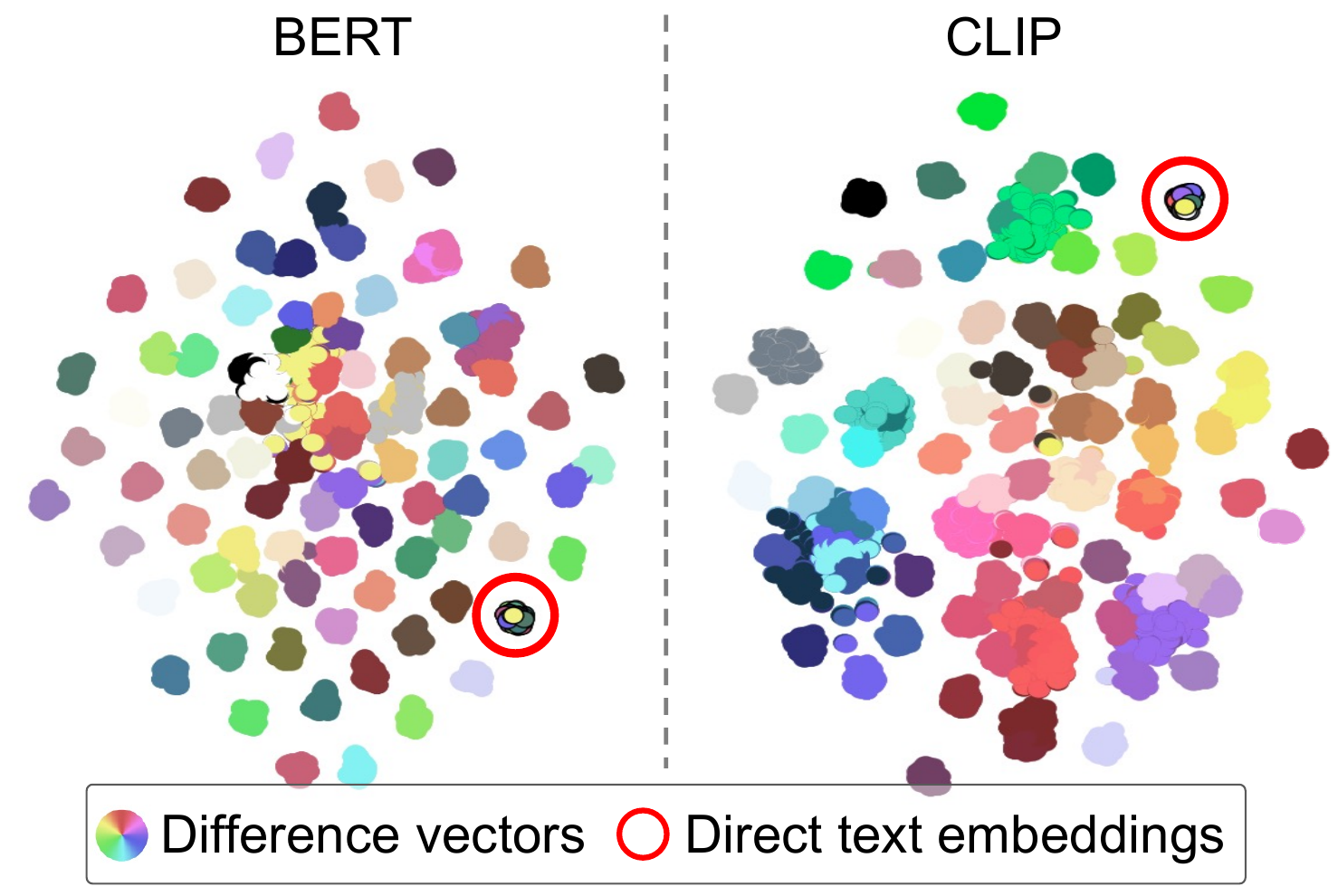}
    % \vspace{-2mm}
    \caption{
    %t-SNE plot of the projected difference vectors, \eg, ``red X'' -- ``X,'' and projected text embeddings (in the red circle), \eg, ``red.''
    %Each point represents a pair of a class in CIFAR-100 and a color attribute.
    The t-SNE plot of difference vectors (\eg, ``brown dog'' -- ``dog'') projected to visual feature space. 
    The colors of the points represent color attributes used for computing the difference vector, and we use all the classes in CIFAR-100 for this plot.
    % visualization. 
    As a comparison, the colored points in the red circle show direct color-text embedding (\eg, ``brown'') projected to the visual feature space.}
    \label{fig:tsne_color}
\end{figure}

%\paragraph{Attribute Embedding vs. Text Embedding}
\paragraph{Embedding Difference vs.~Direct Text Embedding}
When guessing the difference between two
% of
texts, \eg, ``brown X'' -- ``X,'' it would be ``brown.''
Someone may think of using the text embedding directly obtained from ``brown'' instead of our attribute embedding from ``brown X'' -- ``X.''
%To verify the necessity of the difference vector, we visualize the difference  and text embeddings in \Fref{fig:tsne_color}.
To understand the difference between the two representations, we visualize the difference vectors and text embeddings with BERT and CLIP text encoder in \Fref{fig:tsne_color}.
% \oh{(the visualization of other language models can be found in the supplementary material).}
%The t-SNE plot implies that our difference vector is non-trivial in that distinctive color attributes are not represented by the embeddings directly obtained from noun color words, which are in the red circle.
While the direct text embeddings in the red circle of \Fref{fig:tsne_color} are clustered no matter with different color-texts, the difference vectors are well clustered dependent on the color.
This observation indicates that the difference vector is more effective in augmenting the visual feature space than text embedding.
% \son{The text embeddings in the red circle in \Fref{fig:tsne_color} are clustered no matter with different color-texts. 
% The difference vectors, whereas, are clustered dependent on the color.
% used to compute the difference vectors. 
% This observation strongly indicates that the difference vector is more effective way to augment the visual feature space than text embedding.}
In addition, the difference vectors obtained from the same attribute word are similarly clustered regardless of the class ``X'' but slightly different. 
It may imply our attribute embedding has subtle difference awareness on granularity according to class.
% Note that the visualized embeddings are projected ones, and the distributions of the projected and non-projected difference vectors appear similarly on feature space, which means the relationship of texts is properly transferred on the target visual feature space.

Note that \Fref{fig:tsne_color} presents difference vectors in the visual feature space, and we also observe similar distributions of difference vectors in the original text embedding space, \ie, t-SNE is not employed for clustering the direct text embeddings. 
This observation supports our hypothesis that general language models, \eg, BERT or GPT, have learned visual information to some extent.
It, also, demonstrates
% and
the visual information is properly transferred to the target visual feature space.

\noindent\textbf{Do We Need to Rule out Unrealistic Attributes?}
% \moon{for Some Classes}}
One can be curious about how $\TextMani$ handles the unrealistic attribute, such as ``blue cow.''
We intentionally include such unrealistic attributes, motivated by other contexts in 
self- and semi-supervised learning~\cite{chen2021exploring,sohn2020fixmatch,chen2020simple}, where they showed the strong benefit of unnatural strong augmentations to train neural networks.
\moon{This observation regarding strong augmentation is consistent with the design of $\TextMani$ containing unrealistic attributes.}
% In our empirical study, this observation was consistently applied to our case.
% It 
% \son {The observation in the prior works} is consistent with our report of $\TextMani$ containing unrealistic attributes.

% \begin{wraptable}{r}{0.38\linewidth}
%     \centering
%     \vspace{-3mm}
%     \resizebox{0.8\linewidth}{!}{\scriptsize
%     \begin{tabular}{@{\,}l@{\,\,\,}c@{\,}}
%          \toprule
%          \textbf{Aug.} & \textbf{Acc.}\\
%          \midrule
%          Basic       & 31.10\\
%          Random  & 34.04\\
%          \TextMani   & \textbf{34.52}\\
%          \bottomrule
%     \end{tabular}
%     }
%     \caption{Comparison 
%     % of Top-1 accuracy (\%) 
%     to random vector addition on CIFAR-100-10\%.
%     % and class-balanced (CB) one, \ie, CIFAR-100-LT with IF=1. 
%     % with ResNet18.
%     }
%     \label{tab:noise_vec}
%     \vspace{-3mm}
% \end{wraptable}
% \paragraph{Random Words instead of Attributes}
% Our \TextMani\ is an intra-class perturbation, which seems it could be replaced with arbitrary words or noise addition.
% However, \Tref{tab:noise_vec} shows that our difference vector performs favorably to that of random attributes.

\begin{figure}
    \centering
    \includegraphics[width=1.0\linewidth]{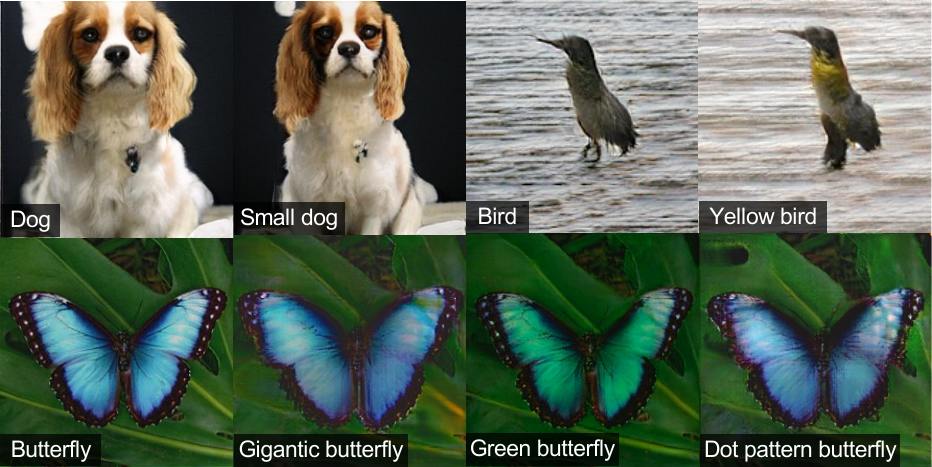}
    \caption{The attribute embedding visualization through image manipulation examples.
    We analyze how an image is manipulated when a difference vector, containing specific attribute information, is injected to the original image feature.
    % An example of images edited using difference vectors. 
    (Top) We visualize an example of generated image given a specific class and its manipulated pair by size and color attribute, respectively.
    (Bottom) From left to right, we visualize the real image of ``Butterfly'' and its manipulated image pair with gigantic, green, and dot pattern, respectively. 
    }
    \label{fig:img_manipulation}
\end{figure}

\noindent\textbf{Does Difference Vector Embed Attribute?}
To visually understand whether attribute editing is reflected while maintaining class information, we attempt to manipulate images by changing their features with the difference vectors $\mathbf{\Delta}_{0\to1} = \bolde_{T_1} {-} \bolde_{T_0}$, \ie, we want to visualize the change effect between $\boldf_{\bI_0}$ and $\bolde_{a_1}= \boldf_{\bI_0} {+} \alpha\mathbf{\Delta}_{0\to1}$ in image domain.
% , so that we can intuitively visualize its effects.
% images manipulated 
To see the effect in image domain, we need to invert the change from $\bolde_{\bI_0}$ to $\bolde_{a_1}$ in image domain, which can be formulated as the following optimization problem, 
\begin{equation}
    \label{eq:l1_loss_naive}
    \begin{aligned}
    % \mathop{\arg\min}_{\theta}\  \| \bolde_{\bI'_0} - \bolde_{a_1} \|_1 = 
    \mathop{\arg\min}\nolimits_{\bI}\  \| E_i(\bI) - \bolde_{a_1} \|_1,
    \end{aligned}
\end{equation}
where $E_i(\cdot): \bI \rightarrow \boldf$ denotes the image encoder in \Fref{fig:textmani}.
Direct optimization in \Eref{eq:l1_loss_naive} is known to be difficult~\cite{zhu2020domain}; thus,
% a difficult optimization~\cite{zhu2020domain}.
we parameterize a given image with an image generator $G_{\theta}$ with a latent code $\bz$, \ie, $\bI(\theta) = G_{\theta}(\bz)$, which is known to ease the optimization~\cite{ulyanov2018deep}. 
Then, we can obtain the visualization by the following optimization over $\theta$
\begin{equation}
    \label{eq:l1_loss}
    \begin{aligned}
    % \mathop{\arg\min}_{\theta}\  \| \bolde_{\bI'_0} - \bolde_{a_1} \|_1 = 
    \mathop{\arg\min}\nolimits_{\theta}\  \| E_i(G_{\theta}(\bz)) - \bolde_{a_1} \|_1,
    \end{aligned}
\end{equation}
where $E_i(\cdot)$, $\bz$, and $\bolde_{a_1}$ are frozen during the optimization.
Since our goal is to see the move from $\boldf_{\bI_0} = E_i(\bI_0)$ to  $\bolde_{a_1}$ for the query $\bI_0$, 
we initialize $\theta$ and $\bz$ such that $G_{\theta}(\bz) \simeq \bI_0$ by the GAN inversion~\cite{zhu2020domain}.
% Note that $\bz$, $E_i(\cdot)$, $\bolde_{a_1}$ the latent vector $\bz$, the encoders $E_i(\cdot)$, and the augmented visual embedding vector $\bolde_{a_1}$ are frozen during the optimization.
% In this work, 
We use IC-GAN~\cite{Perarnau2016icgan} for the image generator and the text embeddings are obtained from the CLIP text encoders. 
Note that $E_i(\cdot)$ is trained with random perturbation of transformed attributes; thus, the visualization through $E_i(\cdot)$ is different from that of CLIP encoders.
Details can be found in the supplementary material.

Figure~\ref{fig:img_manipulation}
% of both real-world and generated images 
shows that the manipulated image reflects the added attribute, \ie,
the size of the dog is reduced by the size attribute ``small,'' and the bird becomes yellow by injecting the color attribute ``yellow.''
% \moon{Also, a single butterfly image becomes various manipulated images reflecting one of the attributes.}
% In addition, there is a result of various attributes reflected in a single butterfly image.
The manipulated results imply that 1) the difference vector indeed embeds the attributes while preserving its semantics, and 2) our augmentation on the feature space may have analogous effects to an image-level augmentation but without implementing complicated image perturbation operations.
Note that these visualizations are for analysis purposes but not for competing with any existing image manipulation methods.

% Difference vector is then measured between the feature vector of $T_0$ and the feature vector of augmented text $T_1$. 
% Using L1 loss, $\bG_{\theta}$ is guided towards generating image that has a feature similar to the augmented vector $\boldf_{a_1}$.The augmented images are shown in \Fref{fig:butterfly_diff}. 
% With pretrained $\bG_{\theta}$ and latent vector $l$, we follow the same scheme as above. The augmented images are shown in \Fref{fig:generator_diff}. 
% The augmented images show a reasonable editing result corresponding to its attribute text. This indicates the residency of semantic meanings in difference vectors.

% Given image As shown in Fig

% \begin{figure}
%     \centering
%     \includegraphics[width=1.0\linewidth]{figures/generator_fig.pdf}
%     \caption{Residual information visualization through generated image manipulation examples.
%     (Left) We visualize an example of generated image given cat class and manipulated image by enormous attribute. 
%     (Right) We show an example of generated butterfly image and an edited image by dark brown color attribute.
%     }
%     \label{fig:generator_diff}
% \end{figure}

%% file: sections/4.experiments.tex
% In this section, we demonstrate the effectiveness of $\TextMani$ in scarce data regimes.
We evaluate $\TextMani$ in various cases presenting sparse data with 
% two
different tasks: long-tail classification in \Sref{sec4.1}, evenly distributed scarce data classification in \Sref{sec4.2}, and few-shot object detection in \Sref{sec4.3}.
We also conduct additional studies demonstrating the effectiveness of the design of our method and the versatility of $\TextMani$
% and ablation study on attributes
% regarding the sampling distribution and the minimum of the
% number of attribute samples for feature manipulation and
% mixing weight $\alpha$ range 
in \Sref{sec4.4}.
Additional experimental results and details can be found in the supplementary material.

\begin{table}
\centering
\resizebox{1.0\linewidth}{!}{
    \footnotesize
    \begin{tabular}{@{}l C{16mm}C{16mm}C{16mm}@{}}
        \toprule
        \multirow{2}[2]{*}{\textbf{(a) Augmentation}} & \multicolumn{3}{c}{\textbf{Imbalance Factor (IF)}} \\ 
        \cmidrule{2-4}
        & \textbf{100} & \textbf{50} & \textbf{10} \\ 
        \midrule
        Baseline           & 38.39 & 43.33 & 59.29 \\
        $\TextMani$ (CLIP) & 40.65 (\blue{+2.26}) & 46.48 (\blue{+3.15}) & 60.17 (\blue{+0.88}) \\ 
        $\TextMani$ (BERT) & 41.10 (\blue{+2.71}) & \textbf{47.17 (\blue{+3.84})} & 60.67 (\blue{+1.38}) \\
        $\TextMani$ (GPT-2) & \textbf{41.20 (\blue{+2.81})} & 46.93 (\blue{+3.60}) & 60.94 (\blue{+1.65}) \\
        \midrule
        Cutout~\cite{devries2017improved}      & 37.51 & 42.28 & 59.26 \\
         + $\TextMani$    & 40.35 (\blue{+2.84}) & 45.48 (\blue{+3.20}) & \textbf{61.31 (\blue{+2.05})} \\
        \cmidrule{1-4}
        Cutmix~\cite{yun2019cutmix}      & 37.93 & 43.34 & 59.30\\
         + $\TextMani$    & 40.22 (\blue{+2.29}) & 45.36 (\blue{+2.02}) & 61.30 (\blue{+2.00})\\
        \cmidrule{1-4}
        Mixup~\cite{zhang2017mixup}       & 36.75 & 40.77 & 57.50 \\
         + $\TextMani$     & 38.40 (\blue{+1.65}) & 43.33 (\blue{+2.56}) & 59.80 (\blue{+2.30})\\
        \cmidrule{1-4}
        ManiMixup~\cite{verma2019manifold}   & 35.72 & 40.51 & 55.26 \\ 
         + $\TextMani$ & 38.60 (\blue{+2.88}) & 43.22 (\blue{+2.71}) & 59.35 (\blue{+4.09})\\
        \midrule
        \midrule
        \multirow{2}[2]{*}{\textbf{(b) Augmentation}} & \multicolumn{3}{c}{\textbf{Set of Classes (IF=100)}} \\
        \cmidrule{2-4}
        & \textbf{Many} & \textbf{Medium} & \textbf{Few}  \\ 
        \midrule
        Baseline           & 71.11 & 38.42 & 3.00 \\
        $\TextMani$ (CLIP) & 71.14 (\blue{+0.03}) & 40.28 (\blue{+1.86}) & 7.53 (\blue{+4.53}) \\
        $\TextMani$ (BERT) & 70.22 (\orange{-0.89}) & 40.73 (\blue{+2.31}) & 9.41 (\blue{+6.41}) \\
        $\TextMani$ (GPT-2) & 70.60 (\orange{-0.51}) & 40.61 (\blue{+2.19}) & \textbf{9.93 (\blue{+6.93})} \\
        \midrule
        Cutout               & 71.54 & 35.94 & 1.06 \\
        + $\TextMani$ & 71.94 (\blue{+0.83}) & \textbf{40.97 (\blue{+2.55})} & 4.03 (\blue{+3.03}) \\
        \cmidrule{1-4}
        Cutmix                & 72.02 & 37.17 & 0.90 \\
         + $\TextMani$  & 72.37 (\blue{+0.35}) & 40.80 (\blue{+3.63}) & 3.90 (\blue{+3.00}) \\
        \cmidrule{1-4}
        Mixup               & 71.97 & 33.62 & 0.36 \\
         + $\TextMani$ & 71.97 (+0.00) & 36.77 (\blue{+3.15}) & 1.83 (\blue{+1.47})\\
        \cmidrule{1-4}
        ManiMixup               & 72.97 & 29.51 & 0.70 \\
         + $\TextMani$ & \textbf{73.20 (\blue{+0.23})} & 36.80 (\blue{+7.29})& 0.76 (\blue{+0.06})\\
        \bottomrule
    \end{tabular}
    }
    \caption{Long-tail classification results (\%) on CIFAR-100-LT with ResNet18.
    (a) The accuracy with respect to the different imbalance factors, \ie, IF=$\{100, 50, 10\}$.
    (b) The accuracy of each class set with IF=$100$.
    Baseline contains random horizontal flip, random crop and rotation, and normalization, applied in all experiments.
    $\TextMani$ without parenthesis uses CLIP for the text encoder.
    }
    \label{tab:CIFAR-100-LT}
    \vspace{4mm}
% \end{table}
% \begin{table}
    \centering
    \resizebox{1.0\linewidth}{!}{\scriptsize
    \begin{tabular}{@{\,\,}l@{\quad}C{5mm}C{9mm}C{9mm}C{9mm}c@{\,\,}}
         \toprule
         \textbf{Aug.} & \textbf{CBS} & \textbf{All} & \textbf{Many} & \textbf{Medium} & \textbf{Few}\\
         \midrule
         Baseline  &            & 38.39 & 71.11 & 38.42 & 3.00 \\
         Cutmix    & \checkmark & 38.23 & \textbf{71.77} & 37.79 & 1.90 \\
         Mixup     & \checkmark & 38.73 & 71.60 & 37.64 & 3.16 \\ 
         ManiMixup & \checkmark & 38.56 & 71.25 & 37.88 & 2.80 \\
         \TextMani & & \textbf{40.65} & 71.14 & \textbf{40.28} & \textbf{7.53} \\
         \bottomrule
    \end{tabular}
    }
    % \vspace{-3mm}
    \caption{Comparison 
    % of Top-1 accuracy (\%) 
    to label mix-based augmentations with class-balanced sampling (CBS) on CIFAR-100-LT with IF=100. 
    % with ResNet18.
    CBS samples two classes first and then samples data in each classes.
    }
    \label{tab:balanced_sampling}
    % \vspace{-3mm}
\end{table}

% textmani for all(Many,mid,few) groups, 30 sampling
% \begin{table}
% \centering
% \resizebox{1.0\linewidth}{!}{
%     \footnotesize
%     \begin{tabular}{@{}ll C{11mm}C{11mm}C{11mm}C{11mm}@{}}
%         \toprule
%         \multirow{2}[2]{*}{\textbf{Method}} &\multirow{2}[2]{*}{\textbf{Aug.}} & \multicolumn{3}{c}{\textbf{Set of Classes}} & \multirow{2}[2]{*}{\textbf{All}} \\ 
%         \cmidrule{3-5}
%         && \textbf{Many} & \textbf{Medium} & \textbf{Few} & \\ 
%         \midrule
%         \multirow{2}{*}{LWS~\cite{kang2019decoupling}}
%           & Baseline    & \textbf{67.65} & 37.52 & 6.03 & 44.84\\
%           & $\TextMani$ & 67.00 & \textbf{39.32} & \textbf{9.49} & \textbf{45.92}\\
%         \midrule 
%         \multirow{2}{*}{cRT~\cite{kang2019decoupling}}
%           & Baseline    & \textbf{67.58} & 38.74 & 9.30 & 45.84 \\ 
%           & $\TextMani$ & 66.91 & \textbf{40.86} & \textbf{13.01} & \textbf{47.10} \\
%           % & $\TextMani$ & 63.61 & \textbf{47.30} & \textbf{24.71} & \textbf{50.50} \\
%         \bottomrule
%     \end{tabular}
% }
% \caption{Long-tail classification accuracy (\%) on ImageNet-LT with ResNext50.
% % LWS represents weight scale learning method, and cRT is a decoupled classifier learning method.
% Baseline stands for the basic augmentation with random horizontal flip and random resized crop.
% % The results show the impact of each augmentation in terms of class sets.
% % \textbf{Bold} indicates the best results in each class set.
% % ResNet-18  
% The models are trained with the batch size of 128.
% }
% \label{tab:ImageNet-LT}
% \end{table}

\subsection{Long-tail Classification}\label{sec4.1}
\paragraph{Experimental Setting}
We compare $\TextMani$ with the mix-based augmentations on the CIFAR-100-LT~\cite{cui2019class} and ImageNet-LT~\cite{liu2019large} datasets, where LT stands for long-tailed distribution.
% The long-tail datasets
They are artificially truncated to have a long-tail from each original dataset, CIFAR-100~\cite{krizhevsky2009learning} and ImageNet-2012~\cite{deng2009imagenet}.
Long-tail datasets usually have three sets of classes: Many-shot (more than 100 images), Medium-shot (20-100 images), and Few-shot (less than 20 images).

For CIFAR-100-LT, we 
% can
control the imbalance factor (IF) \cite{chu2020feature} computed as the ratio of samples in the head to tail class, $N_1/N_K$, where $N_k=\left| \mathcal{D}_k\right|$, and $\mathcal{D}_k$ is the set of samples belonging to the class $k\in\{1,\cdots,K\}$.
% $IF = \max(\{N_i\})/\min(\{N_i\})$, where $N_i$ is the number of training samples of the $i$-th class.
A larger value of the IF represents a more severe imbalance in data, which is more challenging.
We evaluate the performance according to different IFs of 100, 50, and 10.
% We use 100, 50, and 10 for the IF, confirming the performance along the IF value.

We utilize ResNet18 as the baseline on CIFAR-100-LT and ResNext50 on ImageNet-LT.
We use the validation set of the original datasets to measure the Top-1 accuracy.
Note that we apply each augmentation on all the samples without carefully selecting a set of classes in \Tref{tab:CIFAR-100-LT}.

\paragraph{Results}
\moon{\Tref{tab:CIFAR-100-LT} presents the long-tail classification results on CIFAR-100-LT,
% of $\TextMani$, mix-based augmentations, and their combination with $\TextMani$
% are in \Tref{tab:CIFAR-100-LT}.
which show consistent improvement with
% when additionally applying
$\TextMani$.
Also, $\TextMani$ with various text encoders achieves analogous improvement trend regardless of the imbalance factor but marginal degradation on Many class of IF=100 when using general language model, BERT and GPT-2.
While the performance gain is from leaking pre-trained language information, it is surprising and a virtue that the language models never exposed to any image can improve the visual recognition performance.
% It implies that while extracted information from the general language model is helpful for Medium or Few classes due to their scarcity, there is 
% In \Tref{tab:CIFAR-100-LT} for CIFAR-100-LT, we compare the effect of $\TextMani$, mix-based augmentations, and their combination with $\TextMani$.
% Our method achieves the largest improvement among the single usage of augmentation methods in all the imbalance factor conditions regardless of the text encoder type.
% The results show that $\TextMani$ is more effective than other mix-based augmentation when the data distribution has a long-tail, 
In comparison to single usage of mix-based augmentations, our method shows higher accuracy
because of uniform effects of $\TextMani$ on samples regardless of class imbalance.
The mix-based methods, on the other hand, sample two data points from the total dataset, where the probability that a tail class sample contributes to a resulting augmented sample is very low. 
Even with class-balanced sampling on mixed-based augmentation in \Tref{tab:balanced_sampling}, $\TextMani$ performs better, further demonstrating our effectiveness.
}
% The result of \Tref{tab:balanced_sampling} shows that our method performs better even with advantage on mix-based augmentation, and verifies our effectiveness.

\begin{table}
\centering
\resizebox{1.0\linewidth}{!}{
    \footnotesize
    \begin{tabular}{@{}l C{11mm}C{11mm}C{11mm}C{11mm}@{}}
        \toprule
        \multirow{2}[2]{*}{\textbf{Augmentation}} & \multicolumn{3}{c}{\textbf{Set of Classes}} &\multirow{2}[2]{*}{\textbf{Total}}\\
        % & \multirow{2}[2]{*}{\textbf{All}} \\ 
        \cmidrule{2-4}
        & \textbf{Many} & \textbf{Medium} & \textbf{Few}  \\ 
        \midrule
        Baseline & 85.34 & 70.47 & 42.80 & 72.24 \\
        $\TextMani$ & \textcolor{CarnationPink}{85.40} & 71.75 & \textcolor{CarnationPink}{48.49}& \textbf{\textcolor{RedViolet}{73.68}}\\ 
        \cmidrule{1-5}
        Cutout~\cite{devries2017improved} & 85.02 & 70.32 & 42.91 & 72.07 \\ 
         + $\TextMani$ & 85.33 & 71.70 & \textbf{\textcolor{RedViolet}{48.54}}& \textcolor{CarnationPink}{73.65}\\
         \cmidrule{1-5}
        Cutmix~\cite{yun2019cutmix} & 84.85 & 69.90 & 35.82 & 70.77 \\
         + $\TextMani$  & 85.30 & \textcolor{CarnationPink}{71.93} & 47.04 & 73.52 \\
        \cmidrule{1-5}
        Mixup~\cite{zhang2017mixup} & 84.96 & 70.27 & 40.20 & 71.63 \\
         + $\TextMani$ & 84.55 & 69.95 & 35.72 & 70.66 \\
        \cmidrule{1-5}
        ManiMixup~\cite{verma2019manifold} & 84.84 & 69.97 & 38.83 & 71.24 \\
         + $\TextMani$ & \textbf{\textcolor{RedViolet}{85.42}} & \textbf{\textcolor{RedViolet}{71.98}} & 46.65 & 73.54 \\
        \bottomrule
    \end{tabular}
}
\caption{
Long-tail classification results (\%) on ImageNet-LT with ViT, and color the value as \textbf{\textcolor{RedViolet}{best}} and \textcolor{CarnationPink}{second best}.
Baseline contains random horizontal flip, random resize crop, color jitter, and normalization, applied in all experiments.
% \textbf{Bold} indicates the best results in each set of classes, and the value in the parentheses for the improvement or degradation compared to the Baseline.
}
\label{tab:ImageNet-LT_vit}
\end{table}

\begin{table}
    \centering
    \resizebox{1.0\linewidth}{!}{\scriptsize
    \begin{tabular}{@{\,}l@{\quad\quad}C{9mm}C{9mm}C{9mm}C{9mm}@{\,}}
         \toprule
         \textbf{Method} & \textbf{Many} & \textbf{Medium} & \textbf{Few} & \textbf{All}\\
         \midrule
         LWS~\cite{kang2019decoupling} & \textbf{\textcolor{RedViolet}{63.34}} & \textcolor{CarnationPink}{48.08} & \textcolor{CarnationPink}{27.19}& \textcolor{CarnationPink}{51.14}  \\
         cRT~\cite{kang2019decoupling} & 61.80 & 46.20 & 27.40 & 49.60 \\
         cRT+\TextMani & \textcolor{CarnationPink}{62.74} & \textbf{\textcolor{RedViolet}{48.60}} & \textbf{\textcolor{RedViolet}{29.67}} & \textbf{\textcolor{RedViolet}{51.47}} \\
         \bottomrule
    \end{tabular}
    }
    % \vspace{-3mm}
    \caption{Long-tail classification results (\%) on ImageNet-LT with ResNext50. 
    We compare with LWS, cRT, and $\TextMani$ on cRT, and color the value as \textbf{\textcolor{RedViolet}{best}} and \textcolor{CarnationPink}{second best}.
    % The highest accuracy is colored with \textcolor{RedViolet}{purple}, the second one is \textcolor{CarnationPink}{pink}, and the third one is \textcolor{gray}{gray}.
    % The models are trained with a batch size of 512.
    }\vspace{3mm}
    \label{tab:ImageNetLT_512}
\end{table}

Particularly in \Tref{tab:CIFAR-100-LT}(b), the mix-based methods have degraded performance in the Medium and Few-shot classes, while our $\TextMani$ improves performance.
% Although $\TextMani$ with BERT and GPT-2 have marginally degraded accuracy on the Many class, increments on other classes are larger, especially in the Few-shot class.
Combining the mix-based methods with $\TextMani$ improves overall performance, but the tendency to sacrifice the Medium and Few-shot classes is the same as before combining.
Additionally, while Cutout has performance degradation due to the information loss~\cite{yun2019cutmix}, it is not affected by skewness due to no mix between inter-classes; thus, the performance is higher than the mix-based one in the long-tailed distribution.

We also evaluate our $\TextMani$ on the large-scale dataset ImageNet-LT. 
In \Tref{tab:ImageNet-LT_vit}, the best and second best results are with $\TextMani$, which demonstrate that our augmentation method is also effective in the large-scale long-tailed data distribution, consistent with the CIFAR-100-LT results in \Tref{tab:CIFAR-100-LT}.
The improvement with $\TextMani$ implies the importance of intra-class perturbation, which can uniformly affect the samples regardless of the skewness of the class distribution.
% While the probability of getting a tail class sample is getting lower when using the mix-based methods, $\TextMani$ uniformly affects the samples regardless of the skewness of the class distribution.

In \Tref{tab:ImageNetLT_512}, we compare with LWS~\cite{kang2019decoupling}, cRT~\cite{kang2019decoupling}, and $\TextMani$ on cRT. 
LWS and cRT are one of effective methods in recent long-tailed recognition.
% This is the same experiment with Table~\textcolor{blue}{3} in the main paper but with a different batch size of 512.
The result shows that $\TextMani$ on cRT achieves the best results compared to the counterparts in all classes except for the Many class, wherefrom ours achieves second best.
% This is consistent with the result in Table~\textcolor{blue}{3} in the main paper regardless of different batch sizes.
Overall, $\TextMani$ improves well-established works, \eg, LWS, and cRT, and it demonstrates the compatibility of our method.

\begin{table}
\centering
\resizebox{1.0\linewidth}{!}{
\footnotesize
        \begin{tabular}{@{\,}l C{16mm}C{16mm} @{\,}} 
        \toprule
        % \multirow{2}[2]{*}{\textbf{Model}} & 
        \multirow{2}[2]{*}{\textbf{Augmentation}} & \multicolumn{2}{c}{\textbf{Acc.}} \\
        \cmidrule{2-3}
        &  \textbf{Top-1} & \textbf{Top-5}\\
        \midrule
        % \multirow{8}{*}{ResNet18}
          Baseline    & 31.10 & 59.14 \\
          Cutout      & 32.03 & 60.53 \\
          Cutmix      & 32.43 & 61.04 \\
          Mixup       & 32.72 & 62.47 \\
          ManiMixup   & 33.74 & 63.29 \\
        %   $\TextMani$ 30sample & 32.89 & 60.77 \\
        %   $\TextMani$ maxsample & \textbf{34.52} & \textbf{65.74} \\
          $\TextMani$ & \textbf{34.52 (\blue{+3.42})} & \textbf{65.74 (\blue{+6.60})} \\
          \midrule
        %  \cmidrule{2-4}
          Cutout + $\TextMani$    & 33.91 (\blue{+2.81}) & 61.58 (\blue{+2.44}) \\
          Cutmix + $\TextMani$    & 35.61 (\blue{+4.51}) & 63.82 (\blue{+4.68}) \\
          Mixup + $\TextMani$     & 37.97 (\blue{+6.87}) & 66.75 (\blue{+7.61}) \\
          ManiMixup + $\TextMani$ & \textbf{38.02 (\blue{+6.92})} & \textbf{67.28 (\blue{+8.14})} \\

        % conditions
        % 10% cifar100
        % vit-tiny-patch16-224
        % randn fuction
        % common
            % attr: color, size
            % epochs 200, schedule=50,100, 150
            % init_lr=0.1, gamma 0.1, 0.1, 0.1
            % momentum=0.9, decay=5e-5
            % base_aug
                % random horizontal flip
                % random crop size=32, padding=2
                % normalize
        % mixup
            % mixup_alpha 1.0
        % cutout
            % cutout 16
        % cutmix
            % mixup_alpha 1.0
            % cutmix prob 0.5
        % textmani
            % 30, max sample
            % 0.1 scale
            % nsample true
        % textmani + alpha
            % 30 sample
            % nsample
            % 0.1 scale

        \bottomrule
    \end{tabular}
    }
    \caption{
    Classification results (\%) on CIFAR-100-10\% with ResNet18.
    Baseline represents random horizontal flip, random crop, and normalization, basically applied in all experiments.
    % \textbf{Bold} stands for the best results among those with the same number of added augmentations except for Basic, and 
    The parentheses stands for the improvement compared to the Baseline.
    }
    \label{tab:cifar100_10}
    \vspace{3mm}
% \end{table}
% \begin{table}
% \centering
\resizebox{0.95\linewidth}{!}{
\footnotesize
        \begin{tabular}{@{\,}l C{16mm}C{16mm} @{\,}} 
        \toprule
        % \multirow{2}[2]{*}{\textbf{Model}} & 
        \multirow{2}[2]{*}{\textbf{Augmentation}} & \multicolumn{2}{c}{\textbf{Acc.}} \\
        \cmidrule{2-3}
        &  \textbf{Top-1} & \textbf{Top-5}\\
        \midrule
          Baseline    & 65.37 & 89.82 \\
          Cutout      & 69.17 & 91.12 \\
          Cutmix      & 69.82 & 91.76 \\
          Mixup       & 67.54 & 90.23 \\
          $\TextMani$ & \textbf{70.81 (\blue{+5.44})} & \textbf{92.37 (\blue{+2.55})} \\
        %   $\TextMani-30sample$ & 70.81 & 92.37 \\
        %   $\TextMani-maxsample$ & \textbf{71.21} & 92.40 \\
          \midrule
          Cutout + $\TextMani$    & 69.71 (\blue{+4.34}) & 91.32 (\blue{+1.50}) \\
          Cutmix + $\TextMani$    & \textbf{71.05 (\blue{+5.68})} & \textbf{92.22 (\blue{+2.40})} \\
          Mixup + $\TextMani$     & 70.56 (\blue{+5.19}) & 91.58 (\blue{+1.76}) \\
        
        % conditions
        % 10% cifar100
        % resnet50
        % randn fuction
        % common
            % pretrained
            % interpolation 224x224
            % attr: color, size
            % epochs 200, schedule=50,100, 150
            % init_lr=0.1, gamma 0.1, 0.1, 0.1
            % momentum=0.9, decay=5e-4
            % base_aug
                % random horizontal flip
                % random crop size=32, padding=2
                % normalize
        % manimixup
            % mixup_alpha 2.0
        % mixup
            % mixup_alpha 1.0
        % cutout
            % cutout 16
        % cutmix
            % cutmix prob 0.5
            % mixup alpha 1.0
        % textmani
            % 30, max samples
        % textmani + alpha
            % 30 samples
            % nsample True
            % scale min 0.1

        % \midrule
        % \textbf{ViT-T} & \\
        % \midrule
        %   Vanilla & & \\
        %   TextMani & & \\
        % \midrule
        % \textbf{ViT-S} & \\
        % \midrule
        %   Vanilla & & \\
        %   TextMani & & \\
        \bottomrule
    \end{tabular}
    }
    \caption{Classification results (\%) on CIFAR-100-10\% with VIT-Tiny. %pretrain
    % Input image is interpolated to 224x224, 
    % Basic augmentation contains random horizontal flip, random crop, and normalization, which are Basically applied in all experiments.
    The configuration follows \Tref{tab:cifar100_10}.
    % \textbf{Bold} stands for the best results,
    % among those with the same number of added augmentations except for Basic, 
    The parentheses stands for the improvement compared to the Baseline.
    }
    \label{tab:cifar100-10_preVIT}
\end{table}

\subsection{Evenly Distributed Scarce Data Classification}\label{sec4.2}
\paragraph{Experimental Setting}
For evaluating the effectiveness of $\TextMani$ on the scarce dataset, we use 10\% data of the CIFAR-100~\cite{krizhevsky2009learning} and Tiny-ImageNet~\cite{le2015tiny} datasets, named CIFAR-100-10\% and Tiny-ImageNet-10\%, respectively.
CIFAR-100 has 100 classes with 500 training images per class, but we only use randomly sampled 50 images per class.
% The evaluation set is the same as the CIFAR-100 test set containing 10k images. 
Tiny-ImageNet is a subset of ImageNet-1k~\cite{russakovsky2015imagenet} with 100k images and 200 classes, but we use 10k images (50 images per class) for simulating 
% constructing the 
a small dataset.
\moon{Note that the evaluation set is same with those of the original datasets.}
% The evaluation set is the same as the original Tiny-ImageNet one having 10k images.
% , and the validation metrics are Top-1 and 5 accuracies.

The baseline models of scarce data classification are ResNet18~\cite{he2016deep} and ViT-Tiny~\cite{dosovitskiy2020image}.
Due to the space limit, 
% we present the ResNet18 and Vit-Tiny results in this section. 
% the results of other models and 
% the 
details of training can be found in the supplementary material.

\paragraph{Results}
\moon{We demonstrate the effectiveness of $\TextMani$ compared to mix-based augmentations on evenly distributed scarce datasets.}
% We compare the effectiveness of $\TextMani$ and other augmentation methods on the evenly distributed scarce datasets.
As in \Tref{tab:cifar100_10} for CIFAR-100-10\%, $\TextMani$ \moon{outperforms}
% is more effective than 
other methods when a single augmentation is used.
Furthermore, the effect is amplified when our method and mix-based methods are combined, \moon{with particularly good compatibility with Manifold Mixup.}
% and the compatibility with ManiMixup is particularly good.
The results demonstrate the importance of intra-class semantic perturbation along with inter-class in scarce data settings.
\moon{This tendency is also observed with another baseline architecture in \Tref{tab:cifar100-10_preVIT}, and datasets in \Tref{tab:TinyImagenet_10}, implying that
% which implies
$\TextMani$ is model-agnostic to be applied.}
% This tendency is also observed in both the ViT-Tiny (\Tref{tab:cifar100-10_preVIT}) and ResNet18 (\Tref{tab:TinyImagenet_10}) cases.
% evaluated on the Tiny-ImageNet-10\% dataset.
% The results also imply that $\TextMani$ is model-agnostic to be applied.
\moon{The overall results demonstrate the potential of $\TextMani$ to enrich the visual feature space using text modalities and develop more accurate and robust models in scarce data regimes.}

\begin{table}
\centering
% \resizebox{0.8\linewidth}{!}{
% epoch 500 : max sample 10, lr 0.2, attribute color+size
% textmani + other augmentation 은 max sample 10보다 30이 훨씬 좋아서 일단 table에는 30개 sampling으로 작성헀습니다.
\footnotesize
\resizebox{1.0\linewidth}{!}{%
        \begin{tabular}{@{\,}l C{16mm}C{16mm} @{\,}} 
        \toprule
        % \multirow{2}[2]{*}{\textbf{Model}} & 
        \multirow{2}[2]{*}{\textbf{Augmentation}} & \multicolumn{2}{c}{\textbf{Acc.}} \\
        \cmidrule{2-3}
         & \textbf{Top-1} & \textbf{Top-5}\\
        \midrule
        % \multirow{8}{*}{ResNet18}
            Baseline    & 25.94 & 50.53\\
            Cutout      & 26.41 & 50.28\\
            Cutmix      & 25.94 & 49.67\\
            Mixup       & 29.34 & \textbf{54.10}\\
            ManiMixup   & 28.43 & 53.25\\
            $\TextMani$ & \textbf{29.37 (\blue{+3.43})} & 52.37 (\blue{+1.84})\\ 
        %   \cmidrule{2-6}
            \midrule
            Cutout + $\TextMani$    & 29.14 (\blue{+3.20})	& 52.60 (\blue{+2.07}) \\
            Cutmix + $\TextMani$    & 29.86 (\blue{+3.92}) & 54.31 (\blue{+3.78}) \\
            Mixup + $\TextMani$     & 31.15 (\blue{+5.21}) & 56.71 (\blue{+6.18}) \\
            ManiMixup + $\TextMani$ & \textbf{32.39 (\blue{+6.35})} & \textbf{58.25 (\blue{+7.72})} \\
        \bottomrule
    \end{tabular}
    }
    \caption{Classification results on Tiny-ImageNet-10\% with ResNet18. 
    % The results imply Top-1 and 5 accuracies (\%) on the test set.
    The configuration follows \Tref{tab:cifar100_10}.
    % \textbf{Bold} represents the best results, and 
    The parentheses represents the improvement compared to the Baseline.
    }
    \label{tab:TinyImagenet_10}
\end{table}

\subsection{Few-shot Object Detection}\label{sec4.3}
\paragraph{Experimental Setting}
We evaluate $\TextMani$ on the PASCAL VOC~\cite{everingham2010pascal} and MS-COCO~\cite{lin2014microsoft} datasets with a few-shot divison following Wang~\etal~\cite{wang2020frustratingly}.
% ~\cite{qiao2021defrcn}.
For VOC, we have three random splits, which have different divisions into 15 base classes and 5 novel classes among the 20 total classes, and $K=1, 2, 3, 5, 10$ objects are sampled from the novel classes.
We utilize the VOC2007 test set for evaluation with AP50 metrics and train with the combination of the VOC2007 and VOC2012 train/val set.
For COCO, the base classes are disjoint with VOC classes while the remaining classes are used as novel classes, and $K=1, 3, 5, 10, 30$ objects are sampled from the novel classes for few-shot fine-tuning.
We use 5k images from the validation set in COCO for evaluation with mAP metrics and the rest for training.

% The model is trained with the base classes first, and then fine-tuned with the novel classes.
% (few-shot object detection; FSOD setting) or with both base and novel classes (Generalized few-shot object detection; G-FSOD setting).
\moon{The baseline~\cite{yan2019meta} is the Faster R-CNN~\cite{ren2015faster} trained with the base classes first and then fine-tuned 
% the model 
with the novel classes.}
% TFA~\cite{wang2020frustratingly} using
% of the FSOD and G-FSOD 
% , denoted as FRCN, 
% which is the standard model for the object detection task.
% and Decoupled Faster R-CNN~\cite{qiao2021defrcn}, which are denoted as FRCN and DeFRCN, respectively.
\moon{$\TextMani$ is applied to the novel class samples during the fine-tuning stage.}
% at each baseline.
Following the 
% As following 
prior studies, all the reported results are averaged over 10 repeated runs.
% All the results are reproduced based on the DeFRCN~\cite{qiao2021defrcn} code.

\begin{table}
\centering
\resizebox{1.0\linewidth}{!}{
\footnotesize
        \begin{tabular}{@{\,}cl ccccc @{\,}} 
        \toprule
        \multirow{2}[2]{*}{\textbf{Split}} & \multirow{2}[2]{*}{\textbf{Aug.}} & \multicolumn{5}{c}{\textbf{$K$- shot}} \\
        \cmidrule{3-7}
        & & \textbf{1} & \textbf{2} & \textbf{3} & \textbf{5} & \textbf{10} \\
        \midrule
        \multirow{3}{*}{All}
        & Baseline     & 12.82 & 16.65 & 20.04 & 20.64 & 23.19 \\
        & \multirow{2}{*}{$\TextMani$} & \textbf{17.74} & \textbf{22.40} & \textbf{23.37} & \textbf{25.09} & \textbf{24.22} \\
        & & \textbf{(\blue{+4.92})} & \textbf{(\blue{+5.75})} & \textbf{(\blue{+3.33})} & \textbf{(\blue{+4.45})} & \textbf{(\blue{+1.03})} \\
        \midrule
        \midrule
        \multirow{3}{*}{1}
        & Baseline     & 15.11 & 18.82 & 22.61 & 21.97 & 23.74 \\
        & \multirow{2}{*}{$\TextMani$} & \textbf{21.94} & \textbf{26.44} & \textbf{23.66} & \textbf{25.88} & \textbf{25.14} \\
        & & \textbf{(\blue{+6.83})} & \textbf{(\blue{+7.62})} & \textbf{(\blue{+1.05})} & \textbf{(\blue{+3.91})} & \textbf{(\blue{+1.40})} \\
        \midrule
        \multirow{3}{*}{2}
        & Baseline     & 10.86 & 14.22 & 18.67 & 19.34 & 22.49 \\
        & \multirow{2}{*}{$\TextMani$} & \textbf{14.64} & \textbf{18.49} & \textbf{23.28} & \textbf{23.06} & \textbf{24.44} \\
        & & \textbf{(\blue{+3.78})} & \textbf{(\blue{+4.27})} & \textbf{(\blue{+4.61})} & \textbf{(\blue{+3.72})} & \textbf{(\blue{+1.95})} \\
        \midrule
        \multirow{3}{*}{3}
        & Baseline     & 12.49 & 16.90 & 18.84 & 20.61 & 23.35 \\
        & \multirow{2}{*}{$\TextMani$} & \textbf{16.65} & \textbf{22.26} & \textbf{23.16} & \textbf{26.33} & \textbf{25.08} \\
        & & \textbf{(\blue{+4.16})} & \textbf{(\blue{+5.36})} & \textbf{(\blue{+4.32})} & \textbf{(\blue{+5.72})} & \textbf{(\blue{+1.73})} \\
        \bottomrule
        
    % gfsod
    % novel AP50 
    % textmani 30 samples
    % scale 0.1
    
    % defrcn vanilla take only 2 repeat yet
    \end{tabular}
    }
    \caption{Few-shot object detection results (AP50) on VOC. % FSOD setting.
    % The second column represents the types of augmentation methods applied.
    % Baseline stands for no augmentation methods applied.
    % \textbf{Bold} indicates the best results in each model, and 
    The value in the parentheses indicates the improvement compared to the Baseline of each split set.
    }
    \label{tab:voc}
\end{table}

\begin{table}
\centering
\resizebox{1.0\linewidth}{!}{
\footnotesize
        \begin{tabular}{@{\,} l C{10mm}C{10mm}C{10mm}C{10mm}C{10mm} @{\,}} 
        \toprule
        % \multirow{2}[2]{*}{\textbf{Model}} &
        \multirow{2}[2]{*}{\textbf{Aug.}} & \multicolumn{5}{c}{\textbf{$K$- shot}} \\
        \cmidrule{2-6}
        & \textbf{1} & \textbf{3} & \textbf{5} & \textbf{10} & \textbf{30} \\
        \midrule
        % \multirow{2}{*}{FRCN}
         Baseline      & 3.43 & 4.66 & 6.10 &  9.11 & 12.78 \\
         \multirow{2}{*}{$\TextMani$}
         & \textbf{5.39} & \textbf{6.47} & \textbf{7.80} & \textbf{10.03} & \textbf{13.60} \\
         & \textbf{(\blue{+1.96})} & \textbf{(\blue{+1.81})} & \textbf{(\blue{+1.70})} & \textbf{(\blue{+0.92})} & \textbf{(\blue{+0.82})} \\
        % \midrule
        % \multirow{2}{*}{DeFRCN}
        % & Vanilla     & 4.63 & 12.31 & 16.06 & 18.56 & 22.43 \\
        % & $\TextMani$ & & & & & \\
        \bottomrule
    \end{tabular}
    }
    \caption{Few-shot object detection results (mAP) on COCO. % FSOD setting. 
    The configuration follows \Tref{tab:voc}.
    % \textbf{Bold} indicates the best results in each model, and 
    % The parentheses indicates the improvement compared to the Baseline.
    }
    \label{tab:coco}
\end{table}

% \begin{table}
% \centering
% \caption{GFSOD experimental results (mAP) on the COCO dataset.
% }
% \resizebox{1.0\linewidth}{!}{
% \footnotesize
%         \begin{tabular}{@{}ll ccccc @{}} 
%         \toprule
%         \multirow{2}[2]{*}{\textbf{Model}} & \multirow{2}[2]{*}{\textbf{Aug.}} & \multicolumn{5}{c}{\textbf{$K$- shot mAP}} \\
%         \cmidrule{3-7}
%         & & \textbf{1} & \textbf{3} & \textbf{5} & \textbf{10} & \textbf{30} \\
%         \midrule
%         \multirow{2}{*}{FRCN}
%         & Vanilla        & 1.17 (15.51) & 2.72 (14.90) & 3.96 (14.74) & 5.26 (15.15) & 7.32 (15.19) \\
%         & $\TextMani$  & & & & & \\
%         \midrule
%         \multirow{2}{*}{DeFRCN}
%         & Vanilla       & 4.49 (23.73) & 10.30 (26.46) & 13.29 (27.67) & 16.42 (29.51) & 20.70 (31.23) \\
%         & $\TextMani$ & & & & & \\
%         \bottomrule
%     \end{tabular}
%     }
%     \label{tab:coco_g}
% \end{table}

\paragraph{Results}
% Our evaluation on the FSOD task validates that $\TextMani$ can also be applied in the detection task.
% , we evaluate our method on 
Note that we apply $\TextMani$ only on the classification head; thus, the quality of the regressed bounding boxes will remain 
% be the
similar 
% the same
as before applying $\TextMani$.
As shown in \Tref{tab:voc} for VOC and \Tref{tab:coco} for COCO, $\TextMani$ improves the AP by improving only the classification accuracy, where the result 
% follows
is in a similar line to
the analysis~\cite{borji2019empirical} that classification error weighs more than localization error.
The improvement is clearer when $K$ is low.
\moon{The results demonstrate the applicability of $\TextMani$ to enhance the classification accuracy of detection models.}

\begin{table}
    \centering
    \resizebox{1.0\linewidth}{!}{\scriptsize
    \begin{tabular}{@{\,}c@{\,\,\,}l@{\,\,}c@{\quad}c@{\quad}c@{\quad}c@{\quad}c@{\quad}c@{\,}}
         \toprule
         & \textbf{Aug.} & \textbf{Many} & \textbf{Medium} & \textbf{Few}& \textbf{IF=100} & \textbf{IF=50} & \textbf{IF=10}\\
         \midrule
         \multirow{3}{*}{(a)}
         & Baseline   & 71.11 & 38.42 & 3.00 & 38.39 & 43.33 & 59.29  \\ 
         & Random     & \textbf{71.37} & 38.55 & 2.90 & 38.43 & 43.28 & 60.39 \\
         & \TextMani  & 70.22 & \textbf{40.73} & \textbf{9.41} & \textbf{41.10} & \textbf{47.17} & \textbf{60.67} \\
         \midrule
         \multirow{2}{*}{(b)}
         & Direct.    & 71.34 & 38.64 & 4.32 & 38.66 & 43.44 & 59.82 \\
         & Concat.    & 68.02 & 35.82 & 5.35 & 36.98 & 42.68 & 59.44 \\
         \bottomrule
    \end{tabular}
    }
    \caption{Comparison to (a) random perturbation, and
    (b) direct text and concatenated embeddings on CIFAR-100-LT. 
    }
    \label{tab:random}
\end{table}

\subsection{Further Analyses}\label{sec4.4}
% In this section, we further demonstrate the compatibility of our $\TextMani$ with the linear-probed model and conduct an ablation study on attributes.

\paragraph{Random Baseline}
In \Tref{tab:random}-{\color{blue}(a)}, we compare our method with the Random baseline.
We randomly sample a vector from a Normal distribution $\mathcal{N}(0,1)$ and use it instead of the difference vector, \ie, augmenting visual features with random perturbations on the same manifold of visual features.

The result shows that the Random baseline improves performance by serving as intra-perturb, but marginal compared to our method considering semantics additionally, which implies that semantic information embedded in the difference vector guides the augmentation more effective direction rather than random.

\paragraph{Effectiveness of Difference Vectors}
While we use the difference vectors by subtracting the embeddings with and without attribute words for \TextMani, there could be another way to extract the attribute information.
In \Tref{tab:random}-{\color{blue}(b)}, we compare with counterparts, direct text embedding (Direct.) and concatenated embeddings (Concat.).
For the Direct method, we use the text embedding computed from the attribute word directly instead of the difference vector.
For the Concat method, we concatenate the text embeddings from with and without attribute words, \eg, [``bull''$\|$``red bull''], and use it instead of the difference vector.

The results show that using difference vector (\TextMani) outperforms using direct text embedding or concatenated embeddings, and imply that remaining contextual information after subtraction plays an important role in doing intra-perturbation in a semantic way.
% We further discuss the reason for the better performance in the following questions.
Although the word ``blue'' can function as both an adjective and a noun, its exact role in a sentence cannot be determined solely based on the word itself.
Our intention of subtraction is for attribute words to act as a modifier in the sentence motivated by word analogy.
When we computed the cosine similarity, embeddings derived directly from ``red'' and those obtained from the difference exhibited low similarity because they \emph{contain different contextual information} despite the same origin of a word.
% Also, the surrounding meaning can be left even after subtraction to some extent; the class label information would be reflected in the difference vector, which makes the subtle differences (L464-465) in Fig.~{\color{blue}4}.
% Also, the results imply that remaining contextual information after subtraction plays an important role in doing intra-perturbation in a semantic way.

% \begin{table}
\begin{wraptable}{r}{0.46\linewidth}
    \centering
    \vspace{-3mm}
    \resizebox{0.85\linewidth}{!}{\scriptsize
    \begin{tabular}{@{\,}l@{\,\,\,}c@{\,}}
         \toprule
         \textbf{Model} & \textbf{LP-Full}\\
         \midrule
         VL-LTR     & 61.04 \\
         +\TextMani & \textbf{61.82} \\
         \bottomrule
    \end{tabular}
    }
    % \vspace{-3mm}
    \caption{\moon{Comparison between the SOTA model with and without $\TextMani$ during linear probing on CIFAR-100.}
    % Comparison between linear-probed SOTA method and applying our method to it on CIFAR-100.
    % with ResNet18.
    }
    \label{tab:sota}
    \vspace{-2mm}
\end{wraptable}
% \end{table}
\paragraph{Linear Probing with Advanced Models}
\moon{Further demonstrating the compatibility of $\TextMani$, we apply our method during linear probing of the model.}
% when linear-probe the model.
In \Tref{tab:sota}, we test VL-LTR~\cite{tian2022vl}, the state-of-the-art model in long-tail classification, on CIFAR-100.
In \Tref{tab:CLIP_baseline}, we use a CLIP image encoder~\cite{radford2021learning} with various architectures as the baseline model and linear-probe the model on both 10\% and full data of CIFAR-100.
The results demonstrate that $\TextMani$ is compatible with linear-probed CLIP and VL-LTR models.

\begin{table}
    \centering
    \resizebox{0.9\linewidth}{!}{\scriptsize
    \begin{tabular}{@{\ \ }l@{\quad}l@{\quad}c@{\quad}c@{\quad}c@{\ \ }}
         \toprule
         \textbf{CLIP Arch.} & \textbf{Aug.} & \textbf{ZS} & \textbf{LP-10\%} & \textbf{LP-Full}\\
         \midrule
         \multirow{2}{*}{ResNet50} 
         & Baseline  & 39.47 & 50.18 & 63.64 \\
         & \TextMani &   -   & \textbf{52.83} & \textbf{64.17} \\
         \midrule
         \multirow{2}{*}{ResNet101} 
         & Baseline  & 45.17 & 57.37 & 68.60 \\
         & \TextMani &   -   & \textbf{59.49} & \textbf{69.12} \\
         \midrule
         \multirow{2}{*}{ViT-B} 
         & Baseline  & 58.21 & 73.30 & \textbf{79.99} \\
         & \TextMani &   -   & \textbf{73.35} & 79.58 \\
         \bottomrule
    \end{tabular}
    }
    % \vspace{-3mm}
    \caption{Classification results (\%) of CLIP with zero-shot (ZS) and linear-probe (LP) on Full and 10\% CIFAR-100. 
    We apply our $\TextMani$ to the linear-probed CLIP.
    % with ResNet18.
    }
    \label{tab:CLIP_baseline}
    % \vspace{-6mm}
\end{table}

\begin{wraptable}{r}{0.37\linewidth}
    \centering
    % \vspace{-2mm}
    \resizebox{0.75\linewidth}{!}{\scriptsize
    \begin{tabular}{@{\,}c@{\,\,\,}c@{\,\,\,}c@{\,}}
         \toprule
         \textbf{Color} & \textbf{Size} & \textbf{Acc.}\\
         \midrule
                    & & 31.10 \\
         \checkmark & & 33.48 \\
         & \checkmark & 33.89 \\ 
         \checkmark & \checkmark & \textbf{34.52} \\
         \bottomrule
    \end{tabular}
    }
    % \vspace{-3mm}
    \caption{Ablation study on the attributes with CIFAR-100-10\%.
    % with ResNet18.
    }
    \label{tab:ablation_attr}
    \vspace{-2mm}
\end{wraptable}
\paragraph{Ablation Study on Attributes}
\moon{In $\TextMani$, we have considered color and size attributes.}
% We consider color and size attributes in experiments of $\TextMani$.
To confirm the effect of each attribute, we conduct an ablation study on attributes in \Tref{tab:ablation_attr}.
The result shows that while each attribute brings non-trivial gain, using both brings more gain.
We believe that there are additional attributes we could use and a more effective method for selecting appropriate attributes, 
% such as prompt suggestion~\cite{pratt2022does}, 
but leave it for future work.

%% file: sections/5.conclusion.tex
To mitigate the scarce data problem in long-tailed data distribution, small dataset, and few-shot cases, we propose a text-driven visual feature manifold
% space manipulation and
augmentation method, $\TextMani$.
Our method densifies around all the given individual visual features by adding a difference vector stem from the text embedding.
While the mix-based augmentations inflict semantic perturbation in an inter-class way by label mixing, $\TextMani$ perturbs the semantic meaning of the visual features at an intra-class level, \ie, having semantic perturbation while 
% with the
maintaining its class.
% label.
The intra-class semantic perturbation is \moon{achieved}
% done
by transferring the attribute-embedded vectors 
% from text embedding 
to visual feature space.

\moon{To scrutinize the design of our estimated attribute embedding, we conduct visualization-based analyses: t-SNE plot and simple manipulation tests.}
% Our $\TextMani$ is designed on 
% % with the foundation of 
% the hypotheses on the difference attribute vectors.
% % : adding attribute words is reflected in the CLIP text embedding, and the difference before and after addition can be expressed as the difference vector.
% We visualize empirical evidence through 
% % the text embedding and the difference vector by 
% the t-SNE plot and simple manipulation tests, which 
% % and the results 
% empirically support the hypotheses.
\moon{The results empirically demonstrate that $\TextMani$ readily enriches the sparse samples with comprehensible manipulation, since the general language models also reflect some extent of visual information.
The experiment on the long-tail classification validates the effectiveness of our method, especially on the highly skewed class distribution.
% , and also empirically demonstrates some extent of visual information is reflected in the language model.
We additionally show the compatibility of $\TextMani$ with other augmentation methods or other models in scarce data cases and during linear probing.}
% We also validate the effectiveness of our method, especially on the long-tailed distribution and compatibility with the mix-based methods in scarce data cases.
% The experiments with multiple model architectures show that $\TextMani$ is not a model-specific method.
% To mitigate the challenge in FSOD, small dataset, and long-tail class distribution due to the scarce samples, we propose $\TextMani$ exploiting the explainable and manipulatable attributes from the text.
% We transfer the attribute embedded vector by adding a difference in text embeddings in the image feature.
% Our visualization of text embedding and simple image manipulation give evidence that the difference vector contains the residual information, attributes.
% $\TextMani$ can readily enrich the sparse sample with comprehensible zero-shot manipulation, and our experiments demonstrate its effectiveness.
In this work, note that we only use color and size as attributes; thus, there would be room 
\moon{for further investigation of other effective attributes.}
% to investigate other effective
% % various
% attributes further.
% suitable for the classes as future work.

\vspace{4mm}
\noindent
\paragraph{Acknowledgment} 
This work was partially supported by
Institute of Information \& communications Technology Planning \& Evaluation (IITP) 
grant funded by the Korea government(MSIT) (No.2021-0-02068, Artificial Intelligence Innovation Hub; No.2022-0-00124, Development of Artificial Intelligence Technology for Self-Improving Competency-Aware Learning Capabilities; No. 2020-0-00004, Development of Previsional Intelligence based on Long-term Visual Memory Network).